\documentclass[10pt,twocolumn,letterpaper]{article}

\usepackage{cvpr}              % To produce the CAMERA-READY version
\usepackage{algorithm}
\usepackage{algpseudocode}
\usepackage{xcolor}
\usepackage[table, dvipsnames]{xcolor}
\usepackage{hhline}
\usepackage{array}
\usepackage{multirow}
\usepackage{enumitem}
\usepackage{wasysym}

\usepackage{pifont}
\newcommand{\cmark}{\ding{51}}
\newcommand{\xmark}{\ding{55}}

\newcolumntype{P}[1]{>{\centering\arraybackslash}p{#1}}
\newcolumntype{M}[1]{>{\centering\arraybackslash}m{#1}}
\newcommand{\PreserveBackslash}[1]{\let\temp=\\#1\let\\=\temp}
\newcolumntype{C}[1]{>{\PreserveBackslash\centering}p{#1}}
\newcolumntype{R}[1]{>{\PreserveBackslash\raggedleft}p{#1}}
\newcolumntype{L}[1]{>{\PreserveBackslash\raggedright}p{#1}}

\usepackage{tikz}
\usetikzlibrary{tikzmark}

\definecolor{cvprblue}{rgb}{0.21,0.49,0.74}
\usepackage[pagebackref,breaklinks,colorlinks,allcolors=cvprblue]{hyperref}

\def\paperID{7517}
\def\confName{CVPR}
\def\confYear{2025}

\title{What Makes a Good Dataset for Knowledge Distillation?}

\author{Logan Frank \hspace{2cm} Jim Davis \\
Department of Computer Science and Engineering \\
Ohio State University \\
{\tt\small \{frank.580, davis.1719\}@osu.edu}
}

\begin{document}
\maketitle
\begin{abstract}
Knowledge distillation (KD) has been a popular and effective method for model compression. One important assumption of KD is that the teacher's original dataset will also be available when training the student. However, in situations such as continual learning and distilling large models trained on company-withheld datasets, having access to the original data may not always be possible. This leads practitioners towards utilizing other sources of supplemental data, which could yield mixed results. One must then ask: ``what makes a good dataset for transferring knowledge from teacher to student?" Many would assume that only real in-domain imagery is viable, but is that the only option? In this work, we explore multiple possible surrogate distillation datasets and demonstrate that many different datasets, even unnatural synthetic imagery, can serve as a suitable alternative in KD. From examining these alternative datasets, we identify and present various criteria describing what makes a good dataset for distillation. Source code is available at \href{https://github.com/osu-cvl/good-kd-dataset}{\tt \small https://github.com/osu-cvl/good-kd-dataset}.
\end{abstract}    
\section{Introduction} \label{sec:introduction}

Knowledge distillation (KD) \cite{Hinton2015a} has achieved wide-spread popularity as a tool for compressing the information stored inside a large pretrained ``teacher" network into a much more compact and efficient ``student" network. The goal of KD is to train the student to mimic the soft-target outputs (or internal features) of the teacher when presented with similar inputs, which has been shown to often produce a better performing student than training on the task directly. One particular advantage of KD over other model compression techniques is its flexibility in the sense that it allows for the teacher and student to be completely different network architectures (\eg, a vision transformer \cite{Dosovitskiy2020a} distilled into a convolutional neural network \cite{He2016a}).

In practice, one would ordinarily employ the same dataset used to train the teacher model when performing distillation. However, this assumption that the original data will be available may not always hold true. For example, KD is frequently performed for continual learning \cite{Goswami2024a} in the form of self-distillation \cite{Furlanello2018a} (\ie, the teacher and student are the exact same network architecture). Here, data may be ``streamed" where model updating occurs incrementally as new batches of data are obtained, without catastrophically forgetting past data. Another example that is becoming increasingly more common is large networks (and their weights) being released, but the data used to train the network are kept as proprietary in-house information by the company that released the model (\eg, CLIP \cite{Radford2021la}, DINOv2 \cite{Oquab2023a}, and the GPT family \cite{Achiam2023a}). 

Having the inability to access the original dataset has lead users to consider other alternatives for supplemental data, which can come from many different sources and has been explored to varying degrees \cite{Fang2021c, Beyer2022a, Chen2019a}. These sources include: 1) real in-domain (ID) examples, 2) real out-of-domain (OOD) examples, and 3) synthetic examples optimized to be ID. However, there is one, perhaps unlikely, alternative not explored in literature: unoptimized unnatural synthetic OOD imagery (\eg, OpenGL shaders \cite{Shreiner2009a}). 

While we suspect most would elect to use real ID examples as surrogate data, we choose to explore the extreme and ask: \textit{``is it possible to distill knowledge with even the most unconventional dataset?"} When answering this question, we ultimately uncover the mystery of what is required from a particular dataset for it to be successful in KD and show that if certain criteria are met, many different datasets can act as reasonable replacements when the original data are missing. Our contributions are summarized as follows:

\begin{enumerate}[noitemsep,nolistsep]
    \item We identify key characteristics of datasets that enable successful knowledge distillation to a student.
    \item We show that with minimal setup, a teacher can be successfully distilled using unnatural synthetic OOD data.
    \item We present a new adversarial attack strategy that can improve knowledge transfer from teacher to student.
    \item Our work is built on standard knowledge distillation, making it easily applicable to many practitioners.
\end{enumerate}

\noindent We begin with a review of related work in Sect.~\ref{sec:related_work}. The various components of our study are described in Sect.~\ref{sec:method}. Lastly, extensive experiments and analysis of our findings are presented in Sect.~\ref{sec:experiments}.

\section{Related Work} \label{sec:related_work}

In recent years, many works have been proposed for standard KD, using surrogate data in KD, and data-free KD.

\noindent \textbf{Knowledge Distillation.} The transferring of knowledge from a large network to a smaller network was introduced in \cite{Bucilua2006a} and was further refined and coined as ``knowledge distillation'' in \cite{Hinton2015a}. The general KD framework outlined in \cite{Hinton2015a} trains a student network to match the temperature-scaled \cite{Guo2017a} soft outputs from a larger teacher network using entropy-based loss functions. Since then, several works have investigated what properties influence the success of KD \cite{Urban2017a, Cho2019a, Beyer2022a} and others have proposed structural improvements to the seminal approach \cite{Romero2015a, Shen2020a, Tian2020a, Tarvainen2017a}. Notably, \cite{Beyer2022a} argued that KD can be viewed as ``function matching'' and showed that applying strong mixup \cite{Zhang2018a} to the input distillation images results in improved student performance. Adversarial examples were shown to provide accuracy improvements for KD in \cite{Heo2019a, Tian2023a} by identifying the decision boundaries in the teacher.

\noindent \textbf{Utilizing Supplemental Data in Knowledge Distillation.} Many works have considered utilizing a surrogate dataset under the constraint that the original data is unavailable. Transferring knowledge using natural OOD imagery was explored in \cite{Beyer2022a}, however they observed a significant performance loss compared to employing ID data. Some of this performance gap is lessened in \cite{Fang2021c}, where they trained a generator network to leverage OOD data for generating ``in-domain" synthetic examples. In \cite{Asano2023a}, random crops of a single large natural image (\eg, a 2500x2000 pixel image of a busy city street) were used to train the student. Web-scraped real images were utilized in \cite{Tang2022a} to combat the absence of the original training data. The tasks of KD and domain adaptation were combined in \cite{Tang2024a, Tang2024b}, where they sought to train a student to recognize the same classes as the teacher but for a completely different domain (\eg, real images to drawings). Synthetic simulation data was also used to distill models for monocular depth estimation in \cite{Hu2024a}. Some works have even considered pretrained generative AI models for synthesizing realistic data for KD \cite{Li2023a}.

\noindent \textbf{Data-Free Knowledge Distillation.} The potential absence of the original training dataset has motivated a series of data-free KD (DFKD) approaches that attempt to transfer knowledge with no physical data. This line of work can be separated into two categories based on whether they utilize a generator network \cite{Chen2019a, Yoo2019a, Choi2020a, Fang2021a, Binici2022a, Patel2023a, Fang2022a, Yu2023a, Tran2024a, Liu2024a} or inherent teacher network statistics \cite{Nayak2019a, Wang2021a, Yin2020a} to synthesize examples that may be beneficial for KD.

\section{Methodology} \label{sec:method}

In this section, we describe the experimental setup of our study. More specifically: 1) the collection of natural and synthetic datasets from different sources, 2) the role of data augmentation on these datasets, and 3) the method for transferring knowledge from teacher to student.

\subsection{Natural Dataset Collection} \label{sec:natural_dataset}

There is an exorbitant amount of natural (real) imagery available in the world. Thus, when the original training data is not accessible during distillation, employing other real datasets as substitutes would be a logical and reasonable choice. However, there is still one important question: \textit{``which one would work best?"} The obvious answer is to select the dataset that shares the most overlap with the original dataset (\ie, ID), but does the data \textit{need} to be ID with respect to the original dataset or can it be OOD?

To explore this, we employ a wide selection of common image classification datasets spanning general purpose to fine-grained/domain-specific usages. These datasets include CIFAR10 \cite{Krizhevsky2009a}, CIFAR100 \cite{Krizhevsky2009a}, Tiny ImageNet \cite{Deng2009a}, and ImageNet \cite{Deng2009a} for general purpose and FGVC-Aircraft \cite{Maji2013a}, Pets \cite{Parkhi2012a}, Food \cite{Bossard2014a}, and EuroSAT \cite{Helber2019a} for fine-grained/domain-specific. As ImageNet can have a large overlap with many of the classes present in the other datasets, we choose to split it into ID and OOD subsets to fully evaluate how the domain of the data influences KD. When a teacher/dataset has complete overlap of its classes with ImageNet (\eg, Pets), we extract the shared classes and consider them as ID, with any classes remaining in ImageNet being considered OOD. There are some datasets (\eg, CIFAR100) that only have a portion of their classes represented in ImageNet, thus it is an incomplete overlap. In those situations we will not create an ID subset, but all non-overlapping classes can still be treated as OOD. Similarly, we do not include an ID subset for Tiny ImageNet as it is already a direct subset of ImageNet.

When distilling with a dataset different from the original, we limit the number of samples used in KD to 50K (\eg, distilling a CIFAR teacher using 50K ImageNet examples). We select examples for the subsets based on the teacher's predictions, which will be described in the next section as it is similar for synthetic imagery.

\subsection{Synthetic Dataset Collection} \label{sec:synthetic_dataset}

In the previous section we asked whether having ID or OOD \textit{real} images mattered. Here, we expand upon that question by asking: \textit{``does the data even need to be real?"} Several DFKD works have already considered synthetic imagery that has been optimized to be more ID with respect to a particular teacher, which can sometimes result in unnatural images \cite{Yin2020a, Yu2023a, Binici2022a}, but we wonder if the optimization is even necessary. Therefore, we seek to investigate if it is possible to transfer knowledge from a pretrained teacher to a freshly initialized student using solely unoptimized, unnatural OOD synthetic imagery. Such imagery could come in forms such as OpenGL shaders \cite{Shreiner2009a}, Leaves \cite{Baradad2021a}, or noise, each with a different process for obtaining samples. 

For constructing a dataset of OpenGL shaders, we leverage procedural image programs from \cite{Baradad2022a} to render several images for each of the available 1089 TwiGL shaders \cite{Twigl2023a}. As some of these shaders produced images that were either constant (\ie, containing all one color such as all black or all white) or simple (\ie, containing only two colors or containing few colored pixels), we removed such examples in order to create an initial set of synthetic images with the most diversity possible. After rendering and filtering all images, we obtain the initial synthetic dataset $\mathcal{D}_S$.

As for Leaves, we synthesize several images containing randomly arranged shapes using the code provided by \cite{Baradad2021a}, with no filtering needed for obtaining $\mathcal{D}_S$. However, note that the images generated in this format are much more simple and less diverse than the OpenGL shaders, containing only basic shapes such as circles, squares, and triangles that are colored randomly and have no texture (\ie, a constant color for all pixels of the shape). 

Lastly, we construct $\mathcal{D}_S$ consisting of noise images by sampling RGB values for every pixel following a normal distribution defined by the original dataset's statistics. For example, we calculated CIFAR10's mean and standard deviation of the red channel to be 0.5071 and 0.2673, respectively. Thus, we draw a red value from $\mathcal{N}(\mu=0.5071,\ \sigma=0.2673)$ for each pixel in the image to be synthesized (and this is repeated for the green and blue channels). 

\begin{figure}[t]
\begin{center}
\begin{subfigure}{.1\textwidth}
\includegraphics[width=1.7cm]{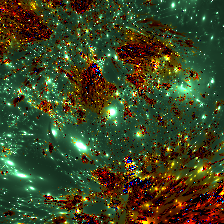}
\end{subfigure}
\begin{subfigure}{.1\textwidth}
\includegraphics[width=1.7cm]{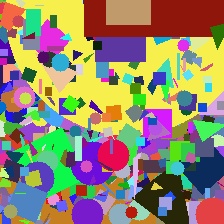}
\end{subfigure}
\begin{subfigure}{.1\textwidth}
\includegraphics[width=1.7cm]{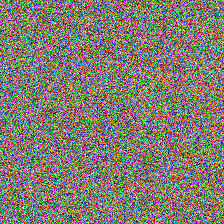}
\end{subfigure}
\\ \vspace{0.1cm}
\begin{subfigure}{.1\textwidth}
\includegraphics[width=1.7cm]{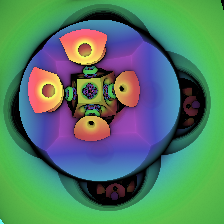}
\end{subfigure}
\begin{subfigure}{.1\textwidth}
\includegraphics[width=1.7cm]{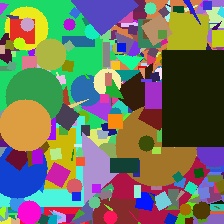}
\end{subfigure}
\begin{subfigure}{.1\textwidth}
\includegraphics[width=1.7cm]{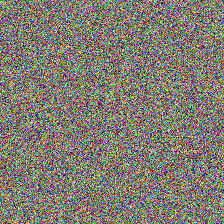}
\end{subfigure}
\end{center}
\vspace{-0.2in}
\caption{Example OpenGL shader (left column), Leaves (middle column), and noise (right column) images.}
\label{fig:synthetic_images}
\vspace{-0.14in}
\end{figure}

As an additional step, we process these initial examples through the teacher network to select a subset to be used for distillation. Given a teacher network $\mathcal{F}_T$ pretrained on some dataset with $\mathcal{C} = \{ c_1, ..., c_R \}$ classes and an initial synthetic dataset $\mathcal{D}_S$, we pass each example in $\mathcal{D}_S$ through $\mathcal{F}_T$ to obtain a teacher prediction and aggregate all predictions. To form the final synthetic dataset $\mathcal{D}_K$ that will be used for KD, we select examples from $\mathcal{D}_S$ based on their teacher predictions. For images predicted as class $c_i$ by the teacher, we randomly sample $N_i$ examples. If a particular class is predicted 0 times, we skip it and sample more examples from the other classes. If a particular class is predicted less than $N_i$ (but more than 0) times, we replicate the examples until we have $N_i$. Collecting $N_i$ synthetic images per class for the specific pretrained teacher creates the base synthetic distillation dataset $\mathcal{D}_K = \{ (x_1, y_1), ..., (x_N, y_N) \}$ where
\begin{align}
    \mathcal{F}_T(x_j) = y_j\ \ \forall\ \ (x_j, y_j) \in \mathcal{D}_K
\end{align}

Examples of these images from possible $\mathcal{D}_S$ sets are shown in Fig.~\ref{fig:synthetic_images}. There is a wide range of complexity/textures and primitive features across the three forms of synthetic imagery and as will be shown in our experiments, this difference in appearance between the OpenGL shaders, Leaves, and noise images will have a pronounced effect on the performance of the distillation.

\subsection{Data Augmentation} \label{sec:data_augmentation}

One particular benefit of the synthetic datasets is the ability to render an unbounded number of images, enabling near-infinite dataset sizes if desired. However, it becomes obvious that obtaining potentially millions of images quickly poses a storage issue. Furthermore, synthesizing new images does not guarantee significantly different examples (from the existing ones), unlike obtaining a new \textit{real} image of some class. A clear solution for increasing dataset diversity without having an over-inflated dataset is data augmentation. This allows us to artificially create more examples during distillation and furthermore, augmentations could enable us to explore regions of the teacher's feature space that the original data could not discover on its own.

In ordinary fully-supervised training, data augmentations should be ``label-preserving'' \cite{Geiping2023a}. However, if we view KD as function matching \cite{Beyer2022a}, then the label-preserving constraint is not an issue. This is especially true in the case of our unnatural synthetic imagery as there is no notion of a ``label" (or any other semantic meaning). Thus, we can employ large amounts of data augmentation that would normally not be considered in standard supervised learning. However, we empirically find (and will show in experiments) that too much data augmentation can harm the final outcome of distillation for real imagery, but not as much for synthetic data. Later, we will describe our complete data augmentation regime for KD and additionally show how data augmentation can provide strong benefits to certain distillation datasets.

\subsection{Knowledge Distillation} \label{sec:knowledge_distillation}

We follow the standard approach for KD \cite{Hinton2015a} in our analysis. Given a pretrained teacher network $\mathcal{F}_T$ and a randomly initialized student network $\mathcal{F}_S$, an example is passed forward through both networks to produce teacher and student softmax distributions $p_T$ and $p_S$, respectively. The KD loss is computed on these softmax scores as
\begin{align}
    \mathcal{L}(p_T\ ||\ p_S) &= \sum_{i\ \in\ \mathcal{C}} \left[\ p_{T(i)} log\ p_{T(i)}\ \text{-}\ p_{T(i)} log\ p_{S(i)}\ \right]
\end{align}

\noindent which is simply the KL-divergence. Like \cite{Hinton2015a}, we also include a temperature parameter $\tau$ to adjust the entropy of the output softmax distributions from the teacher and student networks before they are used to compute the loss.

While more advanced methods of KD do exist, we elect to use the original approach as it is still widely used (and potentially most commonly used) in practice for its simplicity and generality with respect to the network architectures.

\section{Experiments} \label{sec:experiments}

We first conducted experiments performing KD with several different surrogate general purpose, fine-grained/domain-specific, and unnatural synthetic datasets. Next, we investigated what makes certain datasets better for KD than others. We also present a new adversarial attack strategy to explore the importance of having examples close to the decision boundary in KD. Lastly, we compared to existing methods that employ other surrogate data sources.

% \smallskip
\noindent \textbf{Datasets \& Networks.} We employed three general purpose and three fine-grained/domain-specific datasets to train our \textit{teachers}. The general purpose datasets were CIFAR10 (C10), CIFAR100 (C100), and Tiny ImageNet (Tiny) and the fine-grained/domain-specific datasets were FGVC-Aircraft (FGVCA), Pets, and EuroSAT. For our \textit{distillation} datasets, we employed the same six teacher training datasets as well as ImageNet (IN) split into ID and OOD subsets (IN-ID and IN-OOD, respectively), Food, OpenGL shaders, Leaves, and noise. 

In our main distillation experiments, we utilized ResNet50 \cite{He2016a} as the teacher and two different students depending on the teacher's dataset. For CIFAR10/100 trained teachers, we employed ResNet18 and Wide ResNet 40x2 \cite{Zagoruyko2016a} and for all other datasets we employed ResNet18 and MobileNet v2 \cite{Sandler2018a}. In additional experiments, we used stronger teacher models in the form of ConvNeXt-T \cite{Liu2022a} and ViT-S \cite{Touvron2022a}, which were distilled to ResNet18 students.

\noindent \textbf{Teacher Training Details.} To train our ResNet50 teacher models, we used SGD with momentum (0.9) and weight decay (1e-4) and a one-hot cross-entropy loss. All networks were trained for 400 epochs with a batch size of 256 and a half-period cosine learning rate scheduler, beginning with an initial learning rate of 0.1. Data augmentation consisted of RandAugment \cite{Cubuk2020a} ($n=2, m=14$), random horizontal flipping, and random cropping with padding. All ConvNeXt and ViT teachers were finetuned on their respective datasets for 10 epochs using ImageNet1K-pretrained weights, following the settings specified in the ConvNeXt GitHub \cite{Liu2022a}.

% temperature is a hallucination of an example being close to the decision boundary
% really want examples ther eand not faking
% include a one liner on gen AI
% Discussion at the end

\noindent \textbf{Distillation Training Details.} We employed SGD with momentum (0.9) and weight decay (1e-4) for our students, and utilized the standard KD loss as defined previously \cite{Hinton2015a}. Distillation was performed for 400 epochs for all experiments using a half-period cosine learning rate scheduler with an initial learning rate of 0.1, and batch sizes of 256 for all datasets. We empirically found that a temperature value of $\tau = 2$ worked best when distilling with general purpose datasets and $\tau = 20$ for the fine-grained/domain-specific and unnatural synthetic datasets. For real imagery, we utilized the same data augmentation regime as used for training our teacher models. When synthetic imagery was used, we employed stronger data augmentation by increasing the RandAugment amount from $n=2$ to $n=4$ and also added random elastic and random inversion transforms. As previously stated, we utilized stronger data augmentation for the synthetic imagery since there is no notion of a label or any other semantic meaning (compared to real data). However, we did experiment with altering the strength of the data augmentation for both real and synthetic imagery, as will be shown. Furthermore, for both real and synthetic distillation datasets we also included mixup \cite{Zhang2018a} with $\alpha \in \mathcal{U}(0, 1)$, similar to \cite{Beyer2022a}.

As previously mentioned, we capped the number of examples used in distillation at 50K (some datasets may have less because they do not have 50K examples). Thus, we used 50K OpenGL shader, Leaves, and noise images and similarly a 50K subset of the Tiny, IN-ID, IN-OOD, and Food datasets. All other datasets were unchanged (as they have 50K or less examples).

\subsection{Results} \label{sec:results}
We present experimental results of our KD experiments, followed by analyses on the observations from these experiments. We then incorporate our adversarial attack method to enforce certain properties identified in our analyses and conclude with comparisons to other forms of surrogate data.

\begin{table*}[t]
\scriptsize
\setlength\extrarowheight{1pt}
\begin{center}
\begin{tabular}{c || c || c | c | c | c | c | c | c | c | c || c | c | c }
\hline
&  & \multicolumn{9}{c||}{Real} & \multicolumn{3}{c}{Synthetic} \\ 
T$_{\text{DATA}}$  & Student & C10 & C100 & Tiny & FGVCA & Pets & EuroSAT & IN-ID & IN-OOD & Food & OpenGL & Leaves & Noise \\ 
\hhline{=||=||=|=|=|=|=|=|=|=|=||=|=|=}
\multirow{2}{*}{\shortstack[c]{C10\\T: 95.48}} & RN18 & \cellcolor{blue!15} 95.98 & 94.56 & 94.59 & 11.39 & 30.85 & 89.08 & {\bf 94.80} & 94.69 & 93.24 & {\bf 94.02} & 92.08 & 69.24    \\ 
& WRN40x2 & \cellcolor{blue!15} 95.90 & 94.47 & 94.39 & 14.47 & 33.85 & 86.85 & {\bf 94.76} & 94.33 & 91.80 & {\bf 92.59} & 90.22 & 68.89    \\  \hline
 \multirow{2}{*}{\shortstack[c]{C100\\T: 77.45}} & RN18 & 73.98 & \cellcolor{blue!15} 78.35 & 74.71 & 23.01 & 49.11 & 61.59 & - & {\bf 76.66} & 71.45 & {\bf 73.27} & 66.02 & 22.09    \\ 
 & WRN40x2 & 70.17 & \cellcolor{blue!15} 78.01 & 73.08 & 22.14 & 41.76 & 52.03 & - & {\bf 73.63} & 65.34 & {\bf 70.20} & 55.88 & 15.48    \\  \hline
 \multirow{2}{*}{\shortstack[c]{Tiny\\T: 65.66}} & RN18 & 18.64 & 20.69 & \cellcolor{blue!15} 67.14 & 11.31 & 15.67 & 35.08 & - & {\bf 59.44} & 40.08 & {\bf 56.89} & 28.03 & 5.37    \\ 
 & MNv2 & 16.63 & 19.50 & \cellcolor{blue!15} 68.16 & 8.73 & 11.40 & 28.79 & - & {\bf 57.61} & 35.66 & {\bf 55.03} & 24.77 & 2.79   \\  \hhline{=||=||=|=|=|=|=|=|=|=|=||=|=|=}
 \multirow{2}{*}{\shortstack[c]{FGVCA\\T: 91.27}} & RN18 & 11.40 & 10.92 & 38.25 & \cellcolor{blue!15} 89.62 & 16.17 & 7.17 & 71.98 & {\bf 83.91} & 70.54 & {\bf 69.21} & 40.71 & 1.11    \\ 
 & MNv2 & 5.76 & 6.24 & 29.79 & \cellcolor{blue!15} 88.90 & 6.36 & 7.14 & 58.03 & {\bf 75.81} & 54.55 & {\bf 55.66} & 28.92 & 1.08    \\  \hline
 \multirow{2}{*}{\shortstack[c]{Pets\\T: 91.39}} & RN18 &  31.97 & 28.84 & 49.52 & 6.49 & \cellcolor{blue!15} 86.80 & 14.36 & {\bf 90.65} & 85.01 & 60.92 & {\bf 72.59} & 42.19 & 3.43    \\ 
 & MNv2 & 29.92 & 30.34 & 42.19 & 4.03 & \cellcolor{blue!15} 83.7 & 13.65 & {\bf 90.16} & 83.24 & 55.46 & {\bf 67.02} & 37.35 & 3.69    \\  \hline
 \multirow{2}{*}{\shortstack[c]{EuroSAT\\T: 97.80}} & RN18 & 81.20 & 97.55 & 98.25 & 59.00 & 63.65 & \cellcolor{blue!15} 98.60 & - & {\bf 98.55} &  97.55 & {\bf 98.45} & 98.05 & 54.55    \\ 
 & MNv2 & 80.30 & 92.70 & {\bf 98.00} & 51.50 & 66.30 & \cellcolor{blue!15} 98.55 & - &  97.40 &  97.20 & {\bf 98.25} & 97.95 & 51.85    \\  \hline
\end{tabular}
\caption{T$_{\text{DATA}}$ test accuracy of student networks when distilled using various datasets. The models distilled with the original training data (T$_{\text{DATA}}$) are highlighted in purple. The best \textit{surrogate} real and synthetic distillation datasets are emphasized in bold. Entries with a ``-" are those cases where no complete ID subset could be found (as mentioned).}
\vspace{-0.8cm}
\label{tab:main_experiment}
\end{center}
\end{table*}

\noindent \textbf{Standard Knowledge Distillation.} One of the main goals of this study is to uncover what datasets could serve as a surrogate during distillation if the original (teacher) dataset is unavailable. To investigate this, we first na\"ively distilled a ResNet50 teacher (trained on each of the aforementioned teacher datasets) to two different student networks using each of the previously described distillation datasets. These distillation datasets included a mix of general purpose, fine-grained/domain-specific, and synthetic imagery to fully evaluate the wide range of possible alternatives that could be used in KD. Results for each of the teacher training and distillation dataset combinations are shown in Table \ref{tab:main_experiment}.

We see that many different datasets can serve as reasonable substitutes when attempting to transfer knowledge from teacher to student. We dissect our results by answering the questions posed in Sect.~\ref{sec:method}.

\textit{Does the distillation data need to be in-domain?} As seen by the purple cells in Table \ref{tab:main_experiment}, it is clear that distilling with the original dataset often remains superior, however many \textit{real} ID \textit{and} OOD surrogate datasets achieve somewhat comparable results. For CIFAR10, distilling using ImageNet ID and OOD images both come within 1.5\% accuracy of the CIFAR10 distilled student, with ID performing slightly better. The Pets trained teacher is the only scenario where a surrogate actually outperformed the original dataset. However, the number of examples differ drastically, with Pets having roughly 3600 images and ImageNet ID having 50K samples. This is a similar case for FGVCA where the ImageNet OOD dataset outperformed the ID subset, but the OOD version has 50K examples compared to the 3900 images in the ID. With more training time, it is possible that the better performing supplemental datasets (ID or OOD) for Pets and FGVCA could converge and ultimately reach a similar performance level as the teacher (as also suggested in \cite{Beyer2022a}), which could likely be true for CIFAR10/100 and Tiny ImageNet as well. Assuming a student can be trained for long enough, our results suggest that it is possible to transfer knowledge from teacher to student using an alternative OOD dataset. However, we see that having ID data leads to better sample efficiency (\ie, less examples are needed to properly distill the information in the teacher).

\textit{Does the distillation data need to be real?} When we additionally examine the unnatural synthetic distillation datasets, we interestingly observe that a large amount of knowledge can still be transferred to the student for many of the teacher datasets. Distilling using OpenGL shader images obtains within 2\%, 5\%, and 0.2\% of the CIFAR10, CIFAR100, and EuroSAT distilled students, respectively. As for the other synthetic datasets, Leaves is able to come within 4\%, 13\%, and 0.6\% of the CIFAR10, CIFAR100, and EuroSAT distilled students, respectively, but noise does not perform well for any of the teachers (as expected). When the number of classes increases substantially (Tiny ImageNet) or the classes become more fine-grained (FGVCA and Pets), the synthetic images begin to not perform as well, with Leaves and noise struggling significantly more than OpenGL shaders. This could be attributed to needing more training time (similar to the real OOD imagery) or needing more data diversity. 

As shown in Table \ref{tab:main_experiment}, utilizing Leaves over noise achieves large performance gains likely attributed to the inclusion of ``primitive" features (lines and corners, see Fig.~\ref{fig:synthetic_images}) present in the Leaves images. However, OpenGL shaders add even more diversity to the images over Leaves and also includes texture, resulting in substantial performance increases. Since these OpenGL images are rendered from 1089 shader codes using (cyclic) time steps, one could begin to include more shader codes in the synthesis of these images to increase the diversity of the OpenGL shaders dataset, which would likely lead to gains in all scenarios. This is further emphasized with general purpose datasets tending to perform better than the fine-grained ones. Thus, while we show that many teachers can already be distilled using OpenGL shader images, with sufficient training and increased diversity, one could reasonably be able to transfer knowledge to a student using unnatural synthetic imagery (\ie, the data does not need to be real).

\textit{How does the teacher architecture influence what distillation datasets are viable?} We extended our previous experiment for some datasets by examining more powerful teachers (in the form of ConvNeXt-T and ViT-S) to investigate how the choice of teacher model impacts the usability of alternative datasets in KD. These teachers were distilled to ResNet18 models to compare with the previous results. In Table \ref{tab:different_teachers}, we see that the teacher has a significant impact on the \textit{speed} of KD. Not shown in the table, we ran an additional experiment distilling the CIFAR10 and Pets ViT-S teachers to a ResNet18 students using OpenGL shader images for 1200 epochs (as opposed to 400), and saw the performance of the student models increase by over 7\% for both, indicating that ``patient" distillation \cite{Beyer2022a} is especially necessary when the teacher model is more complex and higher performing. Thus, while the choice of teacher certainly impacts the \textit{time} needed to adequately transfer knowledge to a student, it does not appear to impact the \textit{viability} of certain datasets for distillation.

\begin{table}[t]
\scriptsize
\setlength\extrarowheight{1pt}
\begin{center}
\begin{tabular}{c || c | c || c | c | c  }
\hline
T$_{\text{DATA}}$  & \multicolumn{2}{c||}{Teacher} & Original & Food & OpenGL  \\ \hhline{=||==||=|=|=}
\multirow{2}{*}{\shortstack[c]{C10}} & ConvNeXt-T & 98.14 & 97.70 & 56.43 & 87.03    \\ 
 & ViT-S & 98.54 & 97.51 & 72.60 & 83.30    \\  \hline
 \multirow{2}{*}{\shortstack[c]{Tiny}} & ConvNeXt-T & 89.19  & 77.11 & 23.45 & 37.50    \\ 
 & ViT-S & 85.22  & 73.79 & 26.00 & 33.69    \\  \hhline{=||==||=|=|=}
 \multirow{2}{*}{\shortstack[c]{Pets}} & ConvNeXt-T & 94.03  & 86.15 & 38.13 & 32.76    \\ 
& ViT-S & 93.59  & 84.90 & 14.72 & 26.02   \\  \hline
 \multirow{2}{*}{\shortstack[c]{EuroSAT}} & ConvNeXt-T & 98.85  & 98.90 & 95.85 & 97.95    \\ 
& ViT-S & 98.15  & 98.85 & 83.95 & 97.90   \\  \hline
\end{tabular}
\caption{T$_{\text{DATA}}$ test accuracy of a ResNet18 student when distilled from more complex teacher networks.}
\vspace{-0.5cm}
\label{tab:different_teachers}
\end{center}
\end{table}

\noindent \textbf{What Influences Successful Distillation?} We have seen that many different datasets can transfer a large amount of knowledge from teacher to student, but we have also observed that many datasets are not as successful. Thus, what makes a particular dataset more likely to be successful for KD than another? To gain more insight for answering this question, we conducted a series of experiments examining the teacher's predictions, distilling with different levels of teacher output information, and alterations to the distillation dataset and data augmentations. All experiments were performed with a CIFAR10 trained ResNet50 teacher distilled to a ResNet18 student.

\begin{table}[t] 
    \scriptsize
    \setlength\extrarowheight{1pt}
    \begin{center}
    \setlength\tabcolsep{5pt}
    \begin{tabular}{c||c|c|c|c|c|c}
        \hline
        T$_{\text{DATA}}$ & Original & Best & Worst & Food & OpenGL & Leaves \\ \hhline{=||=|=|=|=|=|=}
        C10 & 0.999 & 0.994 & 0.116 & 0.711 & 0.939 & 0.801 \\ \hline
        C100   & 0.999 & 0.983 & 0.631 & 0.834 & 0.918 & 0.808   \\ \hline
        Tiny   & 0.997 & 0.905 & 0.539 & 0.658 & 0.909 & 0.526   \\ \hhline{=||=|=|=|=|=|=}
        FGVCA   & 0.957 & 0.805 & 0.322 & 0.544 & 0.854 & 0.572  \\ \hline
        Pets   & 0.999 & 0.991 & 0.514 & 0.542 & 0.928 & 0.668  \\ \hline
        EuroSAT   & 0.994 & 0.896 & 0.664 & 0.776 & 0.954 & 0.653   \\ \hline
    \end{tabular}
    \caption{Relative entropy of teacher argmax class prediction histograms using various distillation datasets.}
    \vspace{-0.8cm}
    \label{tab:why_entropy}
    \end{center}
\end{table}

In Table \ref{tab:why_entropy}, we computed the relative entropy of the histogram of class counts from the teacher's argmax predictions for a particular distillation dataset (\ie, how many times each class was predicted by the teacher), which we refer to as the class prediction histogram. Relative entropy is computed as $H(p) / H(\mathcal{U}_{\mathcal{C}})$ where $H(\cdot)$ is the entropy function, $p$ is the aforementioned histogram (normalized to sum to 1), and $\mathcal{U}_{\mathcal{C}}$ is the discrete uniform distribution for $\mathcal{C}$ classes (\ie, the distribution that produces the largest entropy). We see that the datasets that perform best in distillation tend to have a relative entropy closer to 1 (the maximum), implying that the teacher predicts all classes equally.

To understand the importance of having a high entropy class prediction histogram (\ie, relative entropy closer to 1), we first changed the teacher's supervisory signal to the student by making the output either one-hot or one-hot with label smoothing \cite{Muller2019a}. We see in Table \ref{tab:why_predictions} that the final scores of the one-hot and label smoothing CIFAR10 students do not differ significantly from the student trained on temperature-scaled softmax outputs (although this is likely to change for more difficult datasets). However, the students distilled with OpenGL shader images benefited more from the temperature-scaled softmax outputs. This could suggest that when distilling using OpenGL data (or any other OOD data), having awareness and understanding of nearby decision boundaries and relationships between classes is especially important (as assisted with temperature scaling).

\begin{table}[t] 
    \scriptsize
    \begin{center}
    \setlength\tabcolsep{5pt}
    \begin{tabular}{c|| P{1.2cm} | P{1.2cm}}
        \hline
        Experiment & C10 & OpenGL  \\ \hhline{=||=|=}
        Original (Temp. Smoothing) & 95.98 & 94.02  \\ \hhline{=||=|=}
        One-Hot   & 96.09 & 91.68     \\ 
        Label Smoothing   & 95.81 & 92.07    \\ \hhline{=||=|=}
        \ \ 20K Long Tail Examples*   & 91.85 & 88.93    \\ 
        20K Long Tail Examples  & 94.36 & 91.39    \\ 
        \ \ 20K Balanced Examples*   &  93.12 & 90.34   \\ 
        20K Balanced Examples   &  95.07 & 92.32   \\ \hline
    \end{tabular}
    \caption{Adjusting the teacher outputs for supervision and altering the distillation dataset based on the teacher's predictions. Experiments with * denote the exclusion of mixup.}
    \vspace{-0.5cm}
    \label{tab:why_predictions}
    \end{center}
\end{table}

To complete our investigation on the importance of having a high entropy class prediction histogram, we reduced the number of examples used in distillation by creating balanced and long tail subsets of 20K examples for both CIFAR10 and OpenGL shaders. As shown in Table \ref{tab:why_predictions}, the long tail version performs about 1.3\% worse than the balanced version (without mixup), indicating that argmax predicting all outputs of the teacher equally is critical for success in KD. However, with mixup we see both experiments perform substantially better, with the gap lessening to about 0.7\%. Thus, being able to explore as much of the teacher's feature space (through mixup) is critical in situations where the number or quality of raw samples is inadequate.

\begin{table}[t] 
    \scriptsize
    \begin{center}
    \setlength\tabcolsep{5pt}
    \begin{tabular}{P{2.5cm}||P{1.2cm}|P{1.2cm}}
        \hline
        Experiment & C10 & OpenGL  \\ \hhline{=||=|=}
        Extreme w/ Mixup & 94.04 & 93.95    \\ 
        High w/ Mixup & 95.69 & \cellcolor{blue!15} {\bf 94.02}  \\
        Standard w/ Mixup   & \cellcolor{blue!15} {\bf 95.98} & 93.39     \\ 
        None w/ Mixup    & 95.00 & 92.20    \\ \hline
        Extreme w/o Mixup    & 93.43 & 92.80    \\ 
        High w/o Mixup    & 95.86 & 93.37    \\ 
        Standard w/o Mixup    &  95.46 & 91.14   \\ 
        None w/o Mixup    &  87.33 & 31.84   \\ \hline
    \end{tabular}
    \caption{CIFAR10 test accuracy of student models when distilled under different levels of data augmentation.}
    \vspace{-0.8cm}
    \label{tab:why_augmentations}
    \end{center}
\end{table}

Lastly, to understand how data diversity plays a role in KD, we adjust the intensity of the data augmentations used during distillation. This includes high and standard (as defined previously), but also adds no data augmentation and ``extreme" data augmentation where we expanded upon the high regime by increasing RandAugment to $n=8,\ m=30$ and added random erasing \cite{Zhong2020a}. We see in Table \ref{tab:why_augmentations} that data augmentation plays a significant role in the performance of the student, but to varying degrees depending on the distillation dataset. From the lowest performing student to the highest, distilling with CIFAR10 gained 8.7\% whereas distilling with OpenGL shaders improved by 62.2\%. Most interestingly, when utilizing extreme augmentations and mixup to distill using CIFAR10, we obtain similar performance as distilling with OpenGL shader data. In other words, when the augmentations becomes too strong, CIFAR10 becomes OOD and we see them perform similarly.

Combining the results from Tables \ref{tab:why_entropy}-\ref{tab:why_augmentations}, we see that KD is a task of function matching \textit{and} sufficient sampling of the teacher. However, not all datasets are equally efficient with their sampling, with ID data demonstrating better sample efficiency than OOD data, and the original data being the most sample efficient of all datasets. One way to understand this is to imagine a signal that needs to be sampled properly for reconstruction (while avoiding aliasing errors). Given a sufficient number of samples, we can better capture what the underlying signal would be similar to how the Nyquist sampling rate operates in signal processing. Thus, having a low entropy class prediction histogram implies that the teacher is being undersampled and its knowledge cannot be reconstructed properly (into the student).

\begin{figure}[t] 
    \scriptsize
    \centering
    % Upper left subfigure
    \begin{subfigure}[b]{0.15\textwidth}
        \centering
        \includegraphics[width=\textwidth]{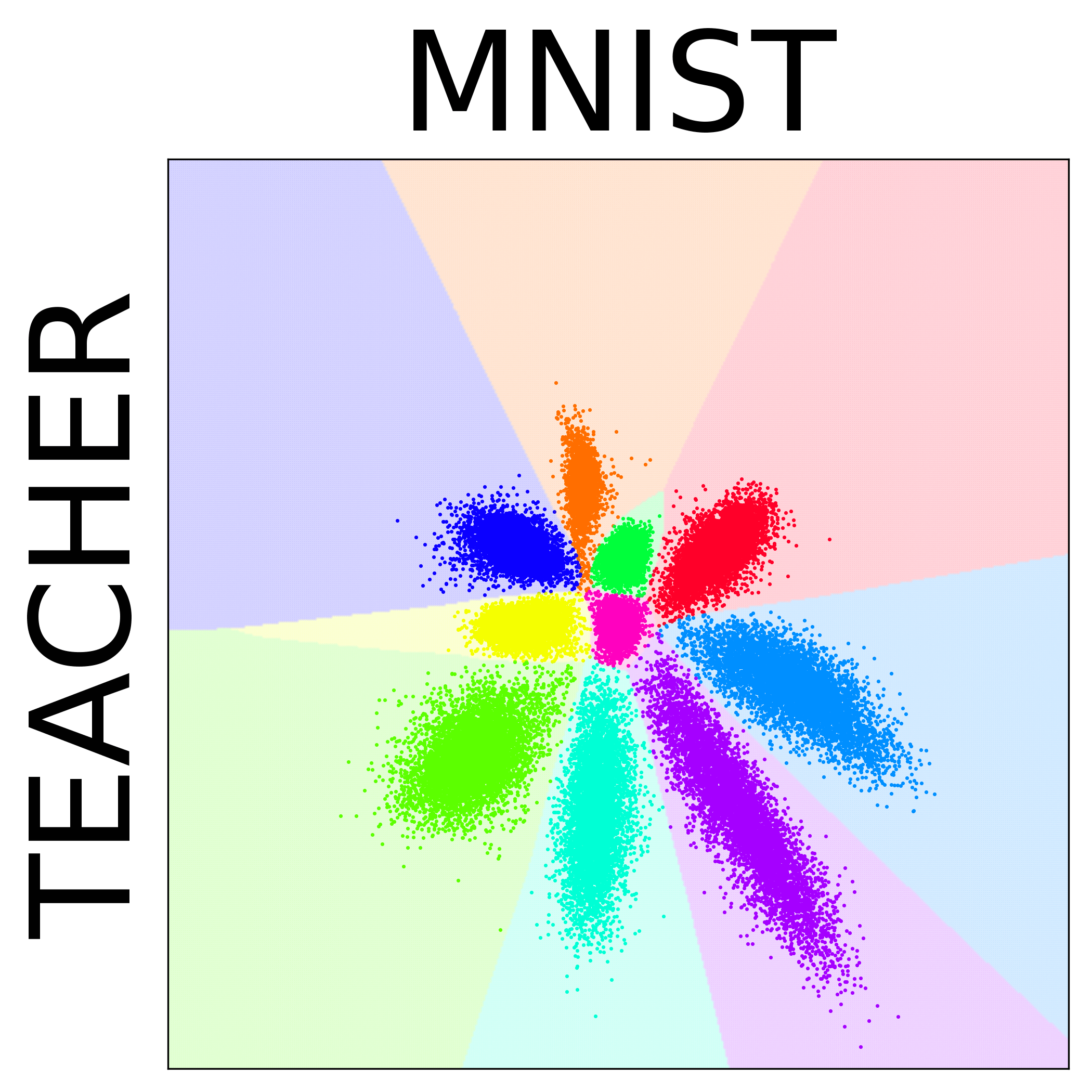}
        \label{fig:tt1}
    \end{subfigure}
    % Upper right subfigure
    \begin{subfigure}[b]{0.15\textwidth}
        \centering
        \includegraphics[width=\textwidth]{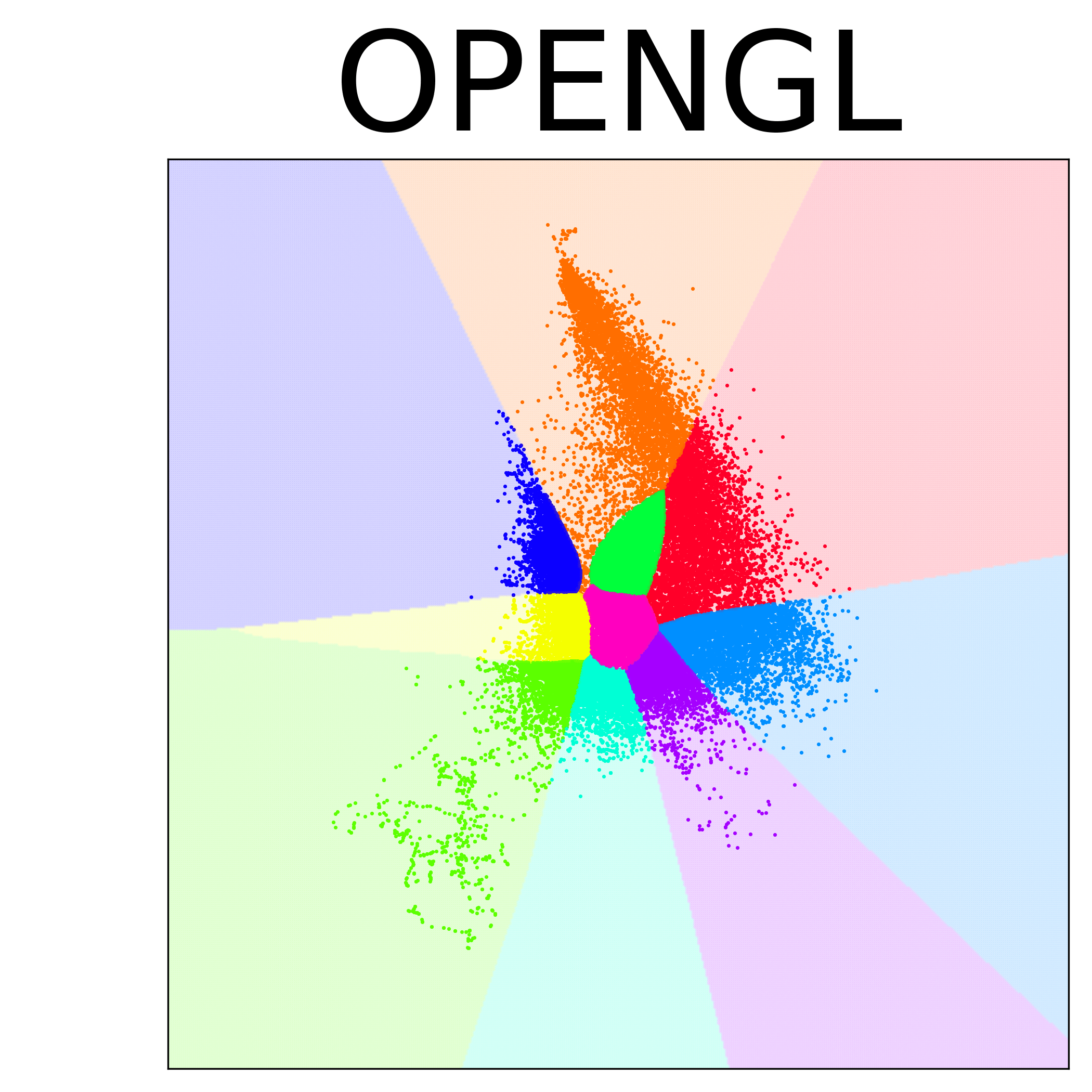} % Replace with your image
        \label{fig:st1}
    \end{subfigure}
    % Upper right subfigure
    \begin{subfigure}[b]{0.15\textwidth}
        \centering
        \includegraphics[width=\textwidth]{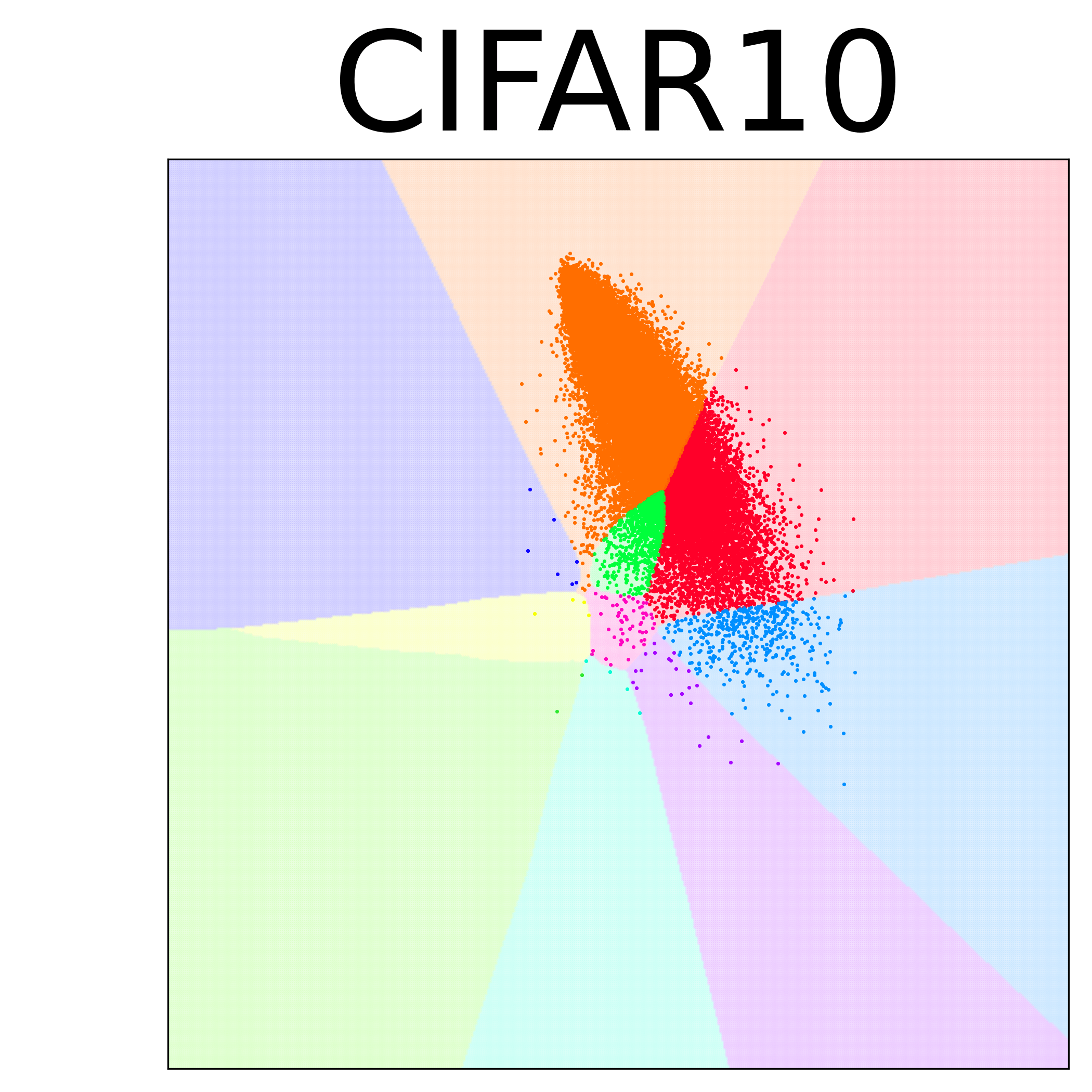} % Replace with your image
        \label{fig:st2}
    \end{subfigure}
    \\ \vspace{-0.25cm}
    % Upper right subfigure
    \begin{subfigure}[b]{0.15\textwidth}
        \centering
        \includegraphics[width=\textwidth, trim=0 0 0 2.5cm, clip]{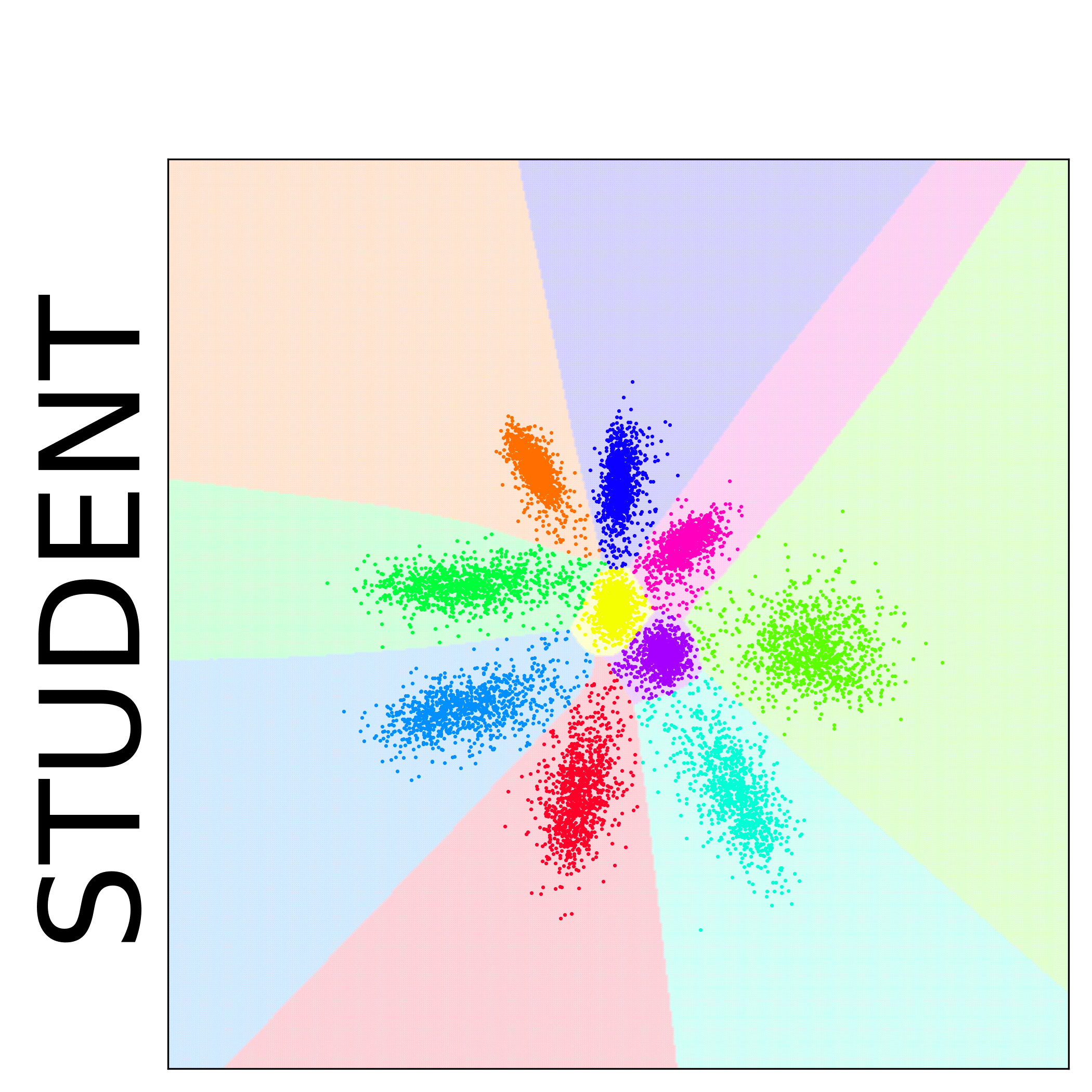} % Replace with your image
        \label{fig:tb1}
    \end{subfigure}
    % Upper right subfigure
    \begin{subfigure}[b]{0.15\textwidth}
        \centering
        \includegraphics[width=\textwidth, trim=0 0 0 2.5cm, clip]{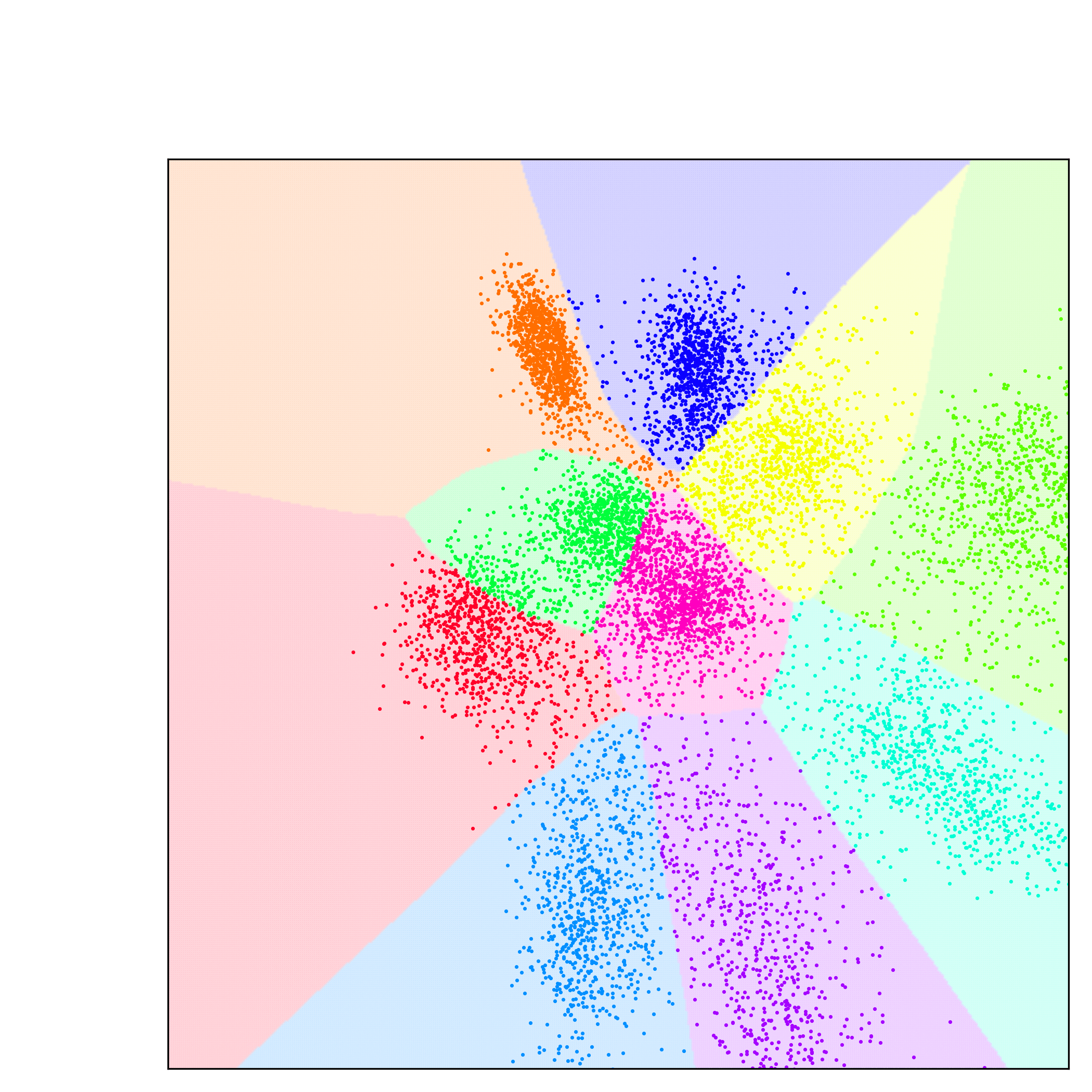} % Replace with your image
        \label{fig:sb1}
    \end{subfigure}
    % Upper right subfigure
    \begin{subfigure}[b]{0.15\textwidth}
        \centering
        \includegraphics[width=\textwidth, trim=0 0 0 2.5cm, clip]{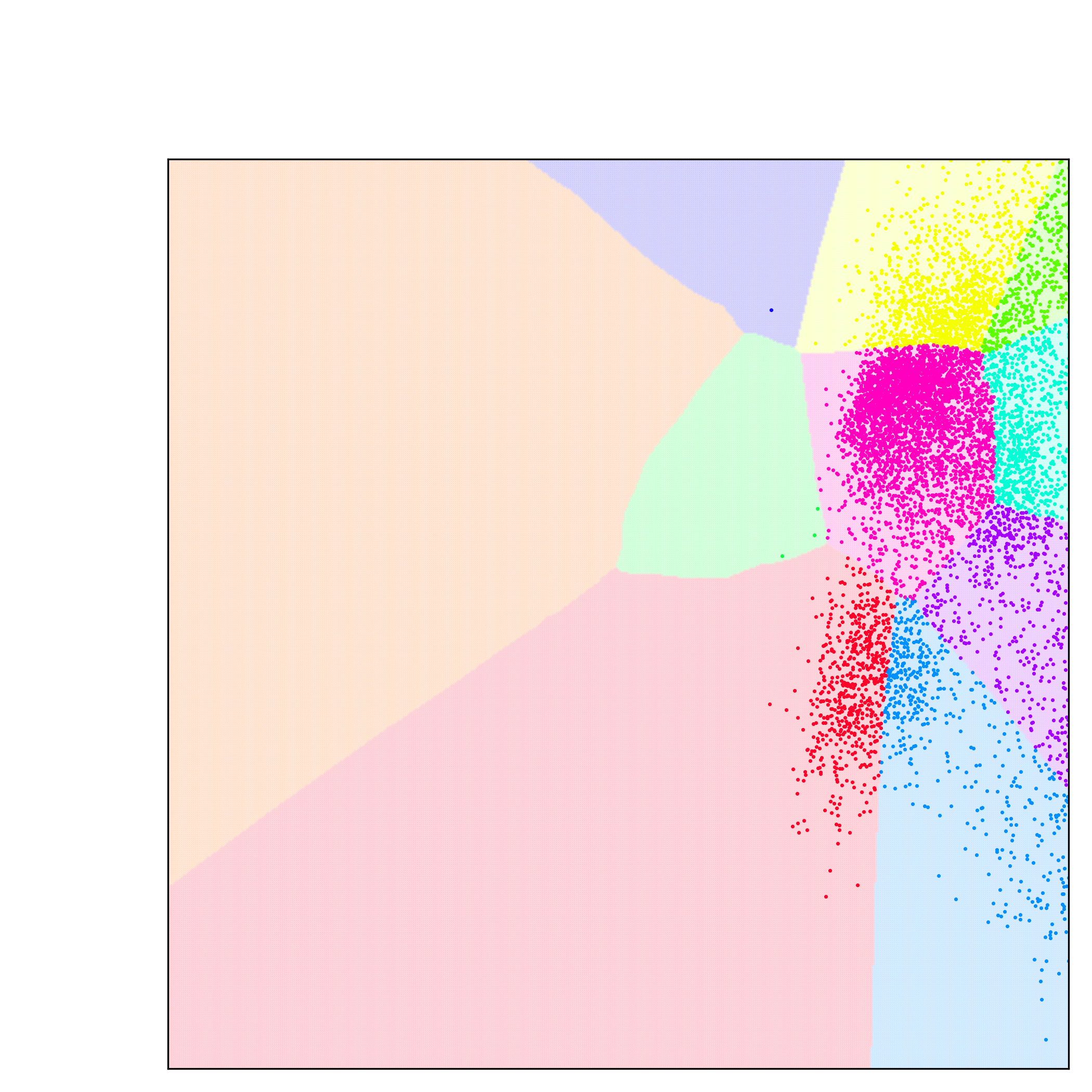} % Replace with your image
        \label{fig:sb2}
    \end{subfigure}
    \vspace{-0.4cm}
    \caption{Visualization of 2D GAP features comparing an MNIST teacher with students trained on various distillation datasets.}
    \vspace{-0.4cm}
    \label{fig:mnist_visualizations}
\end{figure}

To understand the importance of sufficient sampling, we examined a toy scenario by training a simple MNIST teacher (to $\sim$97\% accuracy) that has 3 convolutional layers resulting in 2 GAP features, followed by 3 fully-connected layers (with ReLU activations). We then distilled this teacher model to a student of the exact same architecture using MNIST, OpenGL shader, and CIFAR10 data, then visualized the GAP features of the distillation data through the teacher and the MNIST test data through the distilled students (along with each model's decision boundaries).

From Fig.~\ref{fig:mnist_visualizations}, we see that the OpenGL shader dataset had examples predicted in all class regions of the teacher (resulting in a class prediction histogram with relative entropy closer to 1), while the CIFAR10 data only covered a fraction of the classes (\ie, entropy closer to 0). As a result, the OpenGL shader student obtained decision boundaries more in line with the MNIST student, resulting in an MNIST test accuracy score of 92.89\% compared to 38.78\% accuracy for the CIFAR10 student. Thus, in the case of the OOD experiment in \cite{Beyer2022a}, it is likely that the base OOD argmax predictions in the teacher were imbalanced, which did not sample the decision space in the teacher sufficiently for performing KD, leading to worse knowledge transfer.

\noindent \textbf{Adding Teacher Exploitation.} As we just showed that having a high entropy class prediction histogram plays a significant role in the downstream KD success of a dataset and argued that KD relies on decision boundary information when distilling using surrogate data, one could \textit{force} these properties with adversarial attacks. Thus, if a dataset is deemed ``bad" for KD with a specific teacher, it could be possible that minor perturbations (adversarial attacks) to the images would improve their feasibility.

\begin{figure}[t]
\begin{center}
\begin{subfigure}{.155\textwidth}
\includegraphics[width=2.67cm]{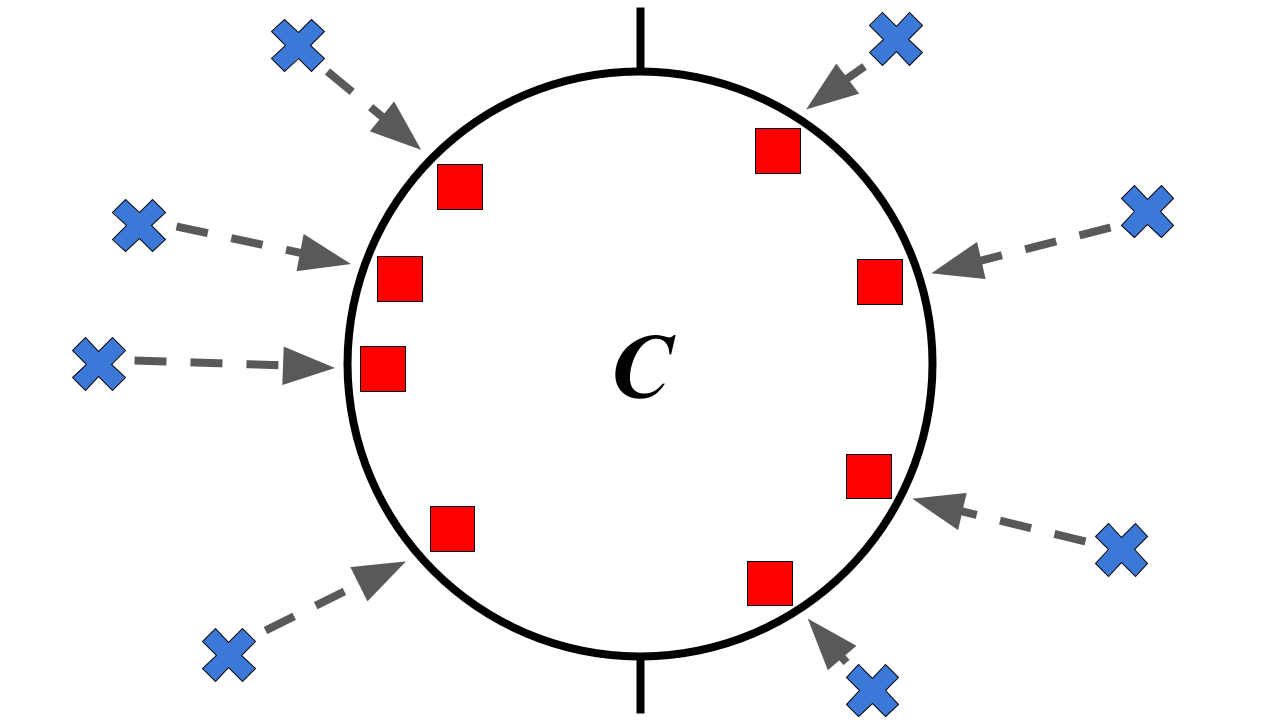}
\caption{}
\end{subfigure} % \\ \vspace{0.3cm}
\begin{subfigure}{.155\textwidth}
\includegraphics[width=2.67cm]{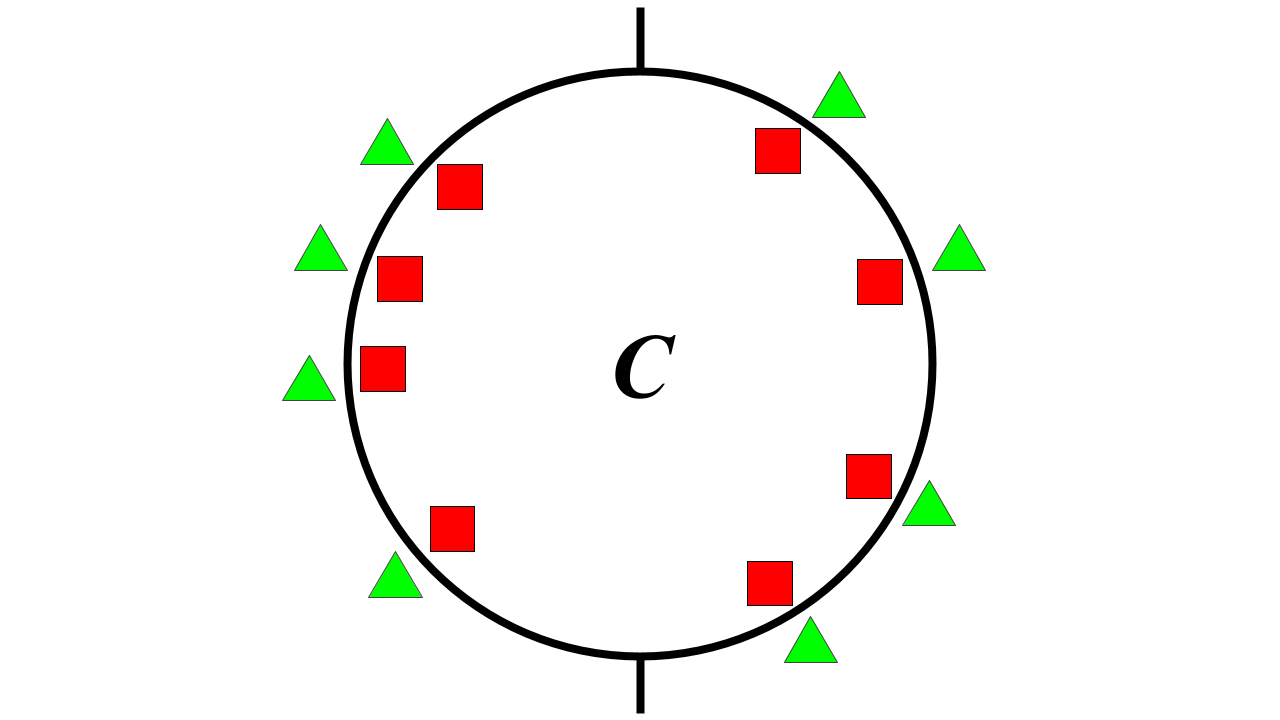}
\caption{}
\end{subfigure}
\begin{subfigure}{.155\textwidth}
\includegraphics[width=2.67cm]{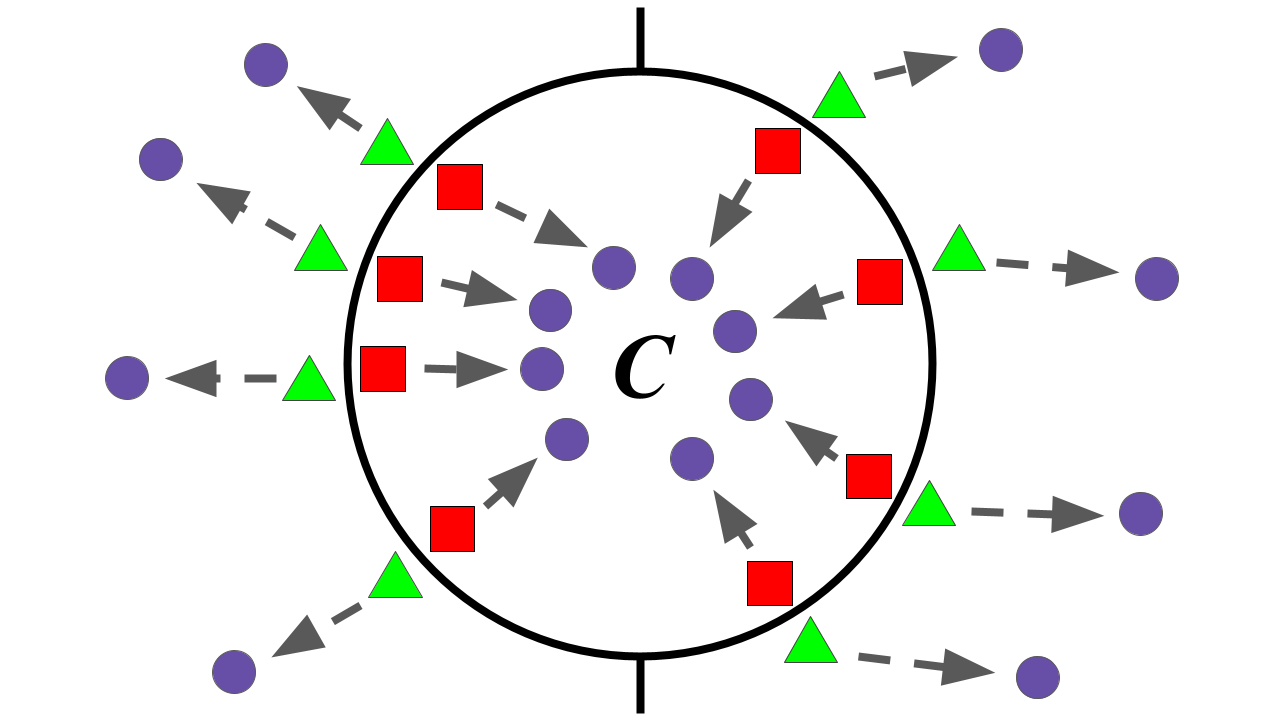}
\caption{}
\end{subfigure}
\end{center}
\vspace{-0.2in}
\caption{Decision boundary exploitation adversarial attack in the teacher feature space for an arbitrary class $C$ (left to right). The symbols \textcolor{blue}{$\times$}, \textcolor{red}{$\blacksquare$}, \textcolor{green}{$\blacktriangle$}, and \textcolor{violet}{$\CIRCLE$} represent the original synthetic examples, ``post''-success examples, ``pre''-success examples, and ``deeper'' examples, respectively. Best viewed in color.}
\vspace{-0.2cm}
\label{fig:adversarial_attack}
\end{figure}

\begin{table}[t]
\scriptsize
\setlength\extrarowheight{1pt}
\begin{center}
\begin{tabular}{c || c || c | c | c | c  }
\hline
T$_{\text{DATA}}$  & Attack & Original & Worst & Food & OpenGL  \\ \hhline{=||=||=|=|=|=}
\multirow{2}{*}{\shortstack[c]{C10\\T: 95.48}} & \cellcolor{red!15} \xmark & {\bf 95.98} & 11.39 & 93.24 & 94.02    \\ 
 & \cellcolor{green!15} \cmark & 95.94 & {\bf 88.14} & {\bf 93.55} & {\bf 94.54}    \\  \hline
 \multirow{2}{*}{\shortstack[c]{C100\\T: 77.45}} & \cellcolor{red!15} \xmark & 78.35 & 23.01 & 71.45 & 73.27    \\ 
 & \cellcolor{green!15} \cmark & {\bf 78.52} & {\bf 58.90} & {\bf 72.64} & {\bf 75.02}    \\  \hhline{=||=||=|=|=|=}
 % \multirow{2}{*}{\shortstack[c]{Tiny\\T: 65.66}} & \cellcolor{red!15} \xmark & 67.14 & 11.31 & {\bf 40.08} & 56.89    \\ 
 % & \cellcolor{green!15} \cmark & {\bf 68.01} & {\bf 15.89} & 39.94 & 00.00    \\  \hhline{=||=||=|=|=|=}
 \multirow{2}{*}{\shortstack[c]{FGVCA\\T: 91.27}} & \cellcolor{red!15} \xmark & 89.62 & 7.17 & 70.54 & 69.21    \\ 
& \cellcolor{green!15} \cmark & {\bf 90.67} & {\bf 10.74} & {\bf 73.60} & {\bf 72.62}   \\  \hline
%  \multirow{2}{*}{\shortstack[c]{Pets\\T: 91.39}} & \cellcolor{red!15} \xmark & 86.80 & 6.49 & 60.92 & 00.00    \\ 
% & \cellcolor{green!15} \cmark & {\bf 88.09} & {\bf 13.67} & 00.00 & 00.00   \\  \hline
 \multirow{2}{*}{\shortstack[c]{EuroSAT\\T: 97.80}} & \cellcolor{red!15} \xmark & 98.60 & 59.00 & 97.55 & 98.45    \\ 
& \cellcolor{green!15} \cmark & {\bf 98.65} & {\bf 96.95} & {\bf 98.45} & {\bf 98.60}   \\  \hline
\end{tabular}
\caption{T$_{\text{DATA}}$ test accuracy of distilled ResNet18 students comparing the inclusion of our adversarial perturbation strategy.}
\vspace{-0.6cm}
\label{tab:attacks}
\end{center}
\end{table}

For our adversarial attack, we perturb an example $x_j$ to an arbitrary label $t$ where $\mathcal{F}_T(x_j) \neq t$ to help identify the decision boundary between $t$ and all other classes. Differing from previous works \cite{Heo2019a, Tian2023a}, we utilize \textit{pairs} of adversarial examples ($x_{1}^{adv} = t,\ x_{2}^{adv} \neq t$) to better outline the decision boundaries (\ie, ``post"-success and ``pre"-success examples) and furthermore require both examples be within some specified argmax softmax threshold to ensure they are somewhat close to the border. We also propose adding a Bold Driver heuristic \cite{Sarkar1995a} to the attack step size to make it more adaptive. Beyond just identifying the decision boundaries in the teacher model, we include an additional single-step attack that copies the pair of examples and perturbs them deeper into the decision space of their respective classes (\ie, more sampling). The full adversarial perturbation process is shown in Fig.~\ref{fig:adversarial_attack} for an arbitrary class $C$. 

By including this adversarial attack to create a more decision boundary aware dataset, we see in Table \ref{tab:attacks} that we can obtain even better performance overall. More interestingly, when applying the attack to the worst performing distillation datasets (from Table \ref{tab:main_experiment}), we observe substantially improved knowledge transfer from teacher to student. In particular, the CIFAR10, CIFAR100, and EuroSAT teachers distilled with FGVCA gained 76.8\%, 35.9\%, and 38\% in accuracy, respectively. However, the FGVCA teacher distilled using EuroSAT imagery does not achieve as significant of gains, likely due to the fine-grained nature of FGVCA. In all scenarios, distilling with the worst dataset using our adversarial perturbation method does not achieve scores as good as Food and OpenGL shaders na\"ively, which is likely due to the limited diversity of imagery present in FGVCA/EuroSAT. The results shown in Table \ref{tab:attacks} further emphasize our points that sufficient sampling of the teacher outputs and diverse imagery are two critical components to the success of a distillation dataset.

\noindent \textbf{Comparisons to Other Data Sources.} As mentioned in Sect.~\ref{sec:related_work}, other methods for creating surrogate KD data exist. Thus, we compared to multiple data-free KD methods \cite{Fang2021a, Binici2022a, Yu2023a, Fang2022a} and a single image KD approach \cite{Asano2023a} using a ResNet34 teacher and ResNet18 student (as commonly done in data-free KD literature). As shown in Table \ref{tab:comparisons}, many of the supplemental datasets perform comparably. Interestingly, most data-free KD methods include loss functions on their generator networks to output all classes equally and to output new examples (\ie, diverse images). However, we find that this can be done directly (as shown with OpenGL shader images) without the need for an additional generator network, which can introduce problems such as mode collapse and non-convergence and furthermore can add substantial computational overhead (Spaceship \cite{Yu2023a} took roughly 48 hours for this experiment compared to 2 hours for vanilla KD).

\begin{table}[t]
\scriptsize
\setlength\extrarowheight{1pt}
\begin{center}
\begin{tabular}{c||c||P{1.2cm}||P{1.2cm}}
\hline
& & \multicolumn{2}{c}{T$_{\text{DATA}}$} \\
Method & S$_{\text{DATA}}$ & C10 & C100 \\ \hhline{=||=||=||=}
 Teacher & - & 95.70 & 78.05 \\ \hline
 Vanilla KD & Original & 95.53 & 79.03 \\ 
 Vanilla KD & Tiny & 94.81 & 76.24 \\ 
 Vanilla KD & Food & 93.20 & 71.74 \\ 
 Vanilla KD & OpenGL & 93.79 & 73.86 \\ \hline
 CMI \cite{Fang2021a} & \multirow{4}{*}{\shortstack[c]{Data-Free}} & 82.40 & 55.20 \\ 
 PRE-DFKD \cite{Binici2022a} & & 87.40 & 70.20 \\ 
 \ \ Spaceship \cite{Yu2023a}* & & 95.19 & 73.40 \\
  FAST \cite{Fang2022a} & & 94.05 & 74.34 \\ \hline
 Single Image \cite{Asano2023a}* & ``City" & 93.47 & 69.34 \\ 
 Single Image \cite{Asano2023a}* & ``Animals" & 93.69 & 71.01 \\
\hline
\end{tabular}
\caption{Comparison of other supplemental data KD approaches on CIFAR10/100 using a ResNet34 teacher and ResNet18 student. Methods with * denote that we reran their code.}
\vspace{-0.8cm}
\label{tab:comparisons}
\end{center}
\end{table}
\section{Discussion} \label{sec:discussion}

This study examined the usage of surrogate datasets in KD when the original data is unavailable, ultimately seeking to answer the question: \textit{``what makes a dataset good for KD?"} Through answering this question, we discovered that the data used for transferring knowledge from teacher to student does \textit{not} need to be ID or real, although ID data (particularly the original data) has the best sample efficiency of all possible distillation datasets. Our results showed that it is entirely possible to perform KD with even the most unconventional datasets, particularly in the form of OpenGL shader images (that did not need to be optimized to a specific teacher). When examining what makes a particular distillation dataset successful as a replacement for the original data in KD, we uncovered that distillation is ultimately a task of sufficient sampling in addition to function matching \cite{Beyer2022a}, with the following being crucial to the success of a distillation dataset:

\begin{itemize}[noitemsep,nolistsep]
    \item All output classes should be represented equally by the data, as measured by the relative entropy of teacher's argmax class prediction histogram.
    \item The data predicted as each individual class should span as much of the decision region as possible (data diversity).
    \item Having greater image diversity/complexity increases the wide-spread applicability and sample-efficiency. 
    \item Increased decision boundary information (softer teacher outputs, physical examples near the border, etc.) is vital in many cases when using surrogate data.
\end{itemize}

\noindent Additionally, we found that increased distillation epochs is especially important when distilling using more complex teachers and when using alternative distillation data. To the best of our knowledge, these collective ideas have not been articulated in literature, with previous works showing that KD with OOD data does not work \cite{Beyer2022a} and others assuming supplemental data is inherently not useful for KD \cite{Fang2021c}.
\section{Conclusion} \label{sec:conclusion}

We investigated the scenario in knowledge distillation where the original data (used to train the teacher) is unavailable for training the student and examined the use of supplemental data. Through our study, we conducted experiments and analyses that give insight into what characteristics indicate that a particular surrogate distillation dataset will better enable knowledge transfer from teacher to student. More specifically, we showed that KD is a sufficient sampling problem that requires the teacher's outputs and decision spaces be equally and thoroughly explored. Our work culminated in experiments showing that it is actually possible to distill many different teacher models using unnatural synthetic imagery in the form of OpenGL shader images. Lastly, we proposed an adversarial perturbation strategy that can improve the knowledge transfer of both well-performing and ineffective distillation datasets.
{
    \small
    \bibliographystyle{ieeenat_fullname}
    \bibliography{main}

\begin{thebibliography}{59}
\providecommand{\natexlab}[1]{#1}
\providecommand{\url}[1]{\texttt{#1}}
\expandafter\ifx\csname urlstyle\endcsname\relax
  \providecommand{\doi}[1]{doi: #1}\else
  \providecommand{\doi}{doi: \begingroup \urlstyle{rm}\Url}\fi

\bibitem[Achiam et~al.(2023)Achiam, Adler, Agarwal, et~al.]{Achiam2023a}
J. Achiam, S. Adler, S. Agarwal, et~al.
\newblock {GPT-4 Technical Report}.
\newblock \emph{arXiv:2303.08774}, 2023.

\bibitem[Asano and Saeed(2023)]{Asano2023a}
Y.~M. Asano and A. Saeed.
\newblock {The Augmented Image Prior: Distilling 1000 Classes by Extrapolating from a Single Image}.
\newblock In \emph{International Conference on Learning Representations}, 2023.

\bibitem[Baradad et~al.(2021)Baradad, Wulff, Wang, Isola, and Torralba]{Baradad2021a}
M. Baradad, J. Wulff, T. Wang, P. Isola, and A. Torralba.
\newblock {Learning to See by Looking at Noise}.
\newblock In \emph{Advances in Neural Information Processing Systems}, 2021.

\bibitem[Baradad et~al.(2022)Baradad, Chen, Wulff, Wang, Feris, Torralba, and Isola]{Baradad2022a}
M. Baradad, R. Chen, J. Wulff, T. Wang, R. Feris, A. Torralba, and P. Isola.
\newblock {Procedural Image Programs for Representation Learning}.
\newblock \emph{Advances in Neural Information Processing Systems}, 2022.

\bibitem[Beyer et~al.(2022)Beyer, Zhai, Royer, Markeeva, Anil, and Kolesnikov]{Beyer2022a}
L. Beyer, X. Zhai, A. Royer, L. Markeeva, R. Anil, and A. Kolesnikov.
\newblock {Knowledge Distillation: A Good Teacher is Patient and Consistent}.
\newblock In \emph{IEEE/CVF Conference on Computer Vision and Pattern Recognition}, 2022.

\bibitem[Binici et~al.(2022)Binici, Aggarwal, Pham, Leman, and Mitra]{Binici2022a}
K. Binici, S. Aggarwal, N.~T. Pham, K. Leman, and T. Mitra.
\newblock {Robust and Resource-Efficient Data-Free Knowledge Distillation by Generative Pseudo Replay}.
\newblock In \emph{AAAI Conference on Artificial Intelligence}, 2022.

\bibitem[Bossard et~al.(2014)Bossard, Guillaumin, and Van~Gool]{Bossard2014a}
L. Bossard, M. Guillaumin, and L. Van~Gool.
\newblock {Food-101 -- Mining Discriminative Components with Random Forests}.
\newblock In \emph{European Conference on Computer Vision}, 2014.

\bibitem[Bucilua et~al.(2006)Bucilua, Caruana, and Niculescu-Mizil]{Bucilua2006a}
C. Bucilua, R. Caruana, and A. Niculescu-Mizil.
\newblock {Model Compression}.
\newblock In \emph{{ACM SIGKDD International Conference on Knowledge Discovery and Data Mining}}, 2006.

\bibitem[Chen et~al.(2019)Chen, Wang, Xu, Yang, Liu, Shi, Xu, Xu, and Tian]{Chen2019a}
H. Chen, Y. Wang, C. Xu, Z. Yang, C. Liu, B. Shi, C. Xu, C. Xu, and Q. Tian.
\newblock {Data-Free Learning of Student Networks}.
\newblock In \emph{IEEE/CVF International Conference on Computer Vision}, 2019.

\bibitem[Cho and Hariharan(2019)]{Cho2019a}
J.~H. Cho and B. Hariharan.
\newblock {On the Efficacy of Knowledge Distillation}.
\newblock In \emph{IEEE/CVF International Conference on Computer Vision}, 2019.

\bibitem[Choi et~al.(2020)Choi, Choi, El-Khamy, and Lee]{Choi2020a}
Y. Choi, J. Choi, M. El-Khamy, and J. Lee.
\newblock {Data-Free Network Quantization with Adversarial Knowledge Distillation}.
\newblock In \emph{IEEE/CVF Conference on Computer Vision and Pattern Recognition Workshops}, 2020.

\bibitem[Cubuk et~al.(2020)Cubuk, Zoph, Shlens, and Le]{Cubuk2020a}
E.~D. Cubuk, B. Zoph, J. Shlens, and Q.~V. Le.
\newblock {RandAugment: Practical Automated Data Augmentation with a Reduced Search Space}.
\newblock In \emph{IEEE/CVF Conference on Computer Vision and Pattern Recognition Workshops}, 2020.

\bibitem[Deng et~al.(2009)Deng, Dong, Socher, Li, Li, and Fei-Fei]{Deng2009a}
J. Deng, W. Dong, R. Socher, L.-J. Li, K. Li, and L. Fei-Fei.
\newblock {ImageNet: A Large-Scale Hierarchical Image Database}.
\newblock In \emph{IEEE Conference on Computer Vision and Pattern Recognition}, 2009.

\bibitem[Dosovitskiy et~al.(2021)Dosovitskiy, Beyer, Kolesnikov, Weissenborn, Zhai, Unterthiner, Dehghani, Minderer, Heigold, Gelly, Uszkoreit, and Houlsby]{Dosovitskiy2020a}
A. Dosovitskiy, L. Beyer, A. Kolesnikov, D. Weissenborn, X. Zhai, T. Unterthiner, M. Dehghani, M. Minderer, G. Heigold, S. Gelly, J. Uszkoreit, and N. Houlsby.
\newblock {An Image is Worth 16x16 Words: Transformers for Image Recognition at Scale}.
\newblock In \emph{International Conference on Learning Representations}, 2021.

\bibitem[Fang et~al.(2021{\natexlab{a}})Fang, Bao, Song, Wang, Xie, Shen, and Song]{Fang2021c}
G. Fang, Y. Bao, J. Song, X. Wang, D. Xie, C. Shen, and M. Song.
\newblock {Mosaicking to Distill: Knowledge Distillation from Out-of-Domain Data}.
\newblock In \emph{Advances in Neural Information Processing Systems}, 2021{\natexlab{a}}.

\bibitem[Fang et~al.(2021{\natexlab{b}})Fang, Song, Wang, Shen, Wang, and Song]{Fang2021a}
G. Fang, J. Song, X. Wang, C. Shen, X. Wang, and M. Song.
\newblock {Contrastive Model Inversion for Data-Free Knolwedge Distillation}.
\newblock In \emph{{International Joint Conference on Artificial Intelligence}}, 2021{\natexlab{b}}.

\bibitem[Fang et~al.(2022)Fang, Mo, Wang, Song, Bei, Zhang, and Song]{Fang2022a}
G. Fang, K. Mo, X. Wang, J. Song, S. Bei, H. Zhang, and M. Song.
\newblock {Up to 100x Faster Data-Free Knowledge Distillation}.
\newblock In \emph{AAAI Conference on Artificial Intelligence}, 2022.

\bibitem[Furlanello et~al.(2018)Furlanello, Lipton, Tschannen, Itti, and Anandkumar]{Furlanello2018a}
T. Furlanello, Z. Lipton, M. Tschannen, L. Itti, and A. Anandkumar.
\newblock {Born Again Neural Networks}.
\newblock In \emph{International Conference on Machine Learning}, 2018.

\bibitem[Geiping et~al.(2023)Geiping, Goldblum, Somepalli, Shwartz-Ziv, Goldstein, and Gordon-Wilson]{Geiping2023a}
J. Geiping, M. Goldblum, G. Somepalli, R. Shwartz-Ziv, T. Goldstein, and A. Gordon-Wilson.
\newblock {How Much Data Are Augmentations Worth? An Investigation into Scaling Laws, Invariance, and Implicit Regularization}.
\newblock In \emph{International Conference on Learning Representations}, 2023.

\bibitem[Goswami et~al.(2024)Goswami, Soutif-Cormerais, Liu, Kamath, Twardowski, and van~de Weijer]{Goswami2024a}
D. Goswami, A. Soutif-Cormerais, Y. Liu, S. Kamath, B. Twardowski, and J. van~de Weijer.
\newblock {Resurrecting Old Classes with New Data for Exemplar-Free Continual Learning}.
\newblock In \emph{IEEE/CVF Conference on Computer Vision and Pattern Recognition}, 2024.

\bibitem[Guo et~al.(2017)Guo, Pleiss, Sun, and Weinberger]{Guo2017a}
C. Guo, G. Pleiss, Y. Sun, and K.~Q. Weinberger.
\newblock {On Calibration of Modern Neural Networks}.
\newblock In \emph{International Conference on Machine Learning}, 2017.

\bibitem[He et~al.(2016)He, Zhang, Ren, and Sun]{He2016a}
K. He, X. Zhang, S. Ren, and J. Sun.
\newblock {Deep Residual Learning for Image Recognition}.
\newblock In \emph{IEEE/CVF Conference on Computer Vision and Pattern Recognition}, 2016.

\bibitem[Helber et~al.(2019)Helber, Bischke, Dengel, and Borth]{Helber2019a}
P. Helber, B. Bischke, A. Dengel, and D. Borth.
\newblock {EuroSAT: A Novel Dataset and Deep Learning Benchmark for Land Use and Land Cover Classification}.
\newblock \emph{{IEEE Journal of Selected Topics in Applied Earth Observations and Remote Sensing}}, 2019.

\bibitem[Heo et~al.(2019)Heo, Lee, Yun, and Choi]{Heo2019a}
B. Heo, M. Lee, S. Yun, and J.~Y. Choi.
\newblock {Knowledge Distillation with Adversarial Samples Supporting Decision Boundary}.
\newblock In \emph{AAAI Conference on Artificial Intelligence}, 2019.

\bibitem[Hinton et~al.(2015)Hinton, Vinyals, and Dean]{Hinton2015a}
G. Hinton, O. Vinyals, and J. Dean.
\newblock {Distilling the Knowledge in a Neural Network}.
\newblock \emph{{arXiv Preprint arXiv:1503.02531}}, 2015.

\bibitem[Hu et~al.(2024)Hu, Fan, Ozay, Jiang, and Lam]{Hu2024a}
J. Hu, C. Fan, M. Ozay, H. Jiang, and T.~L. Lam.
\newblock {Dense Depth Distillation with Out-of-Distribution Simulated Images}.
\newblock \emph{{Knowledge-Based Systems}}, 2024.

\bibitem[Krizhevsky(2009)]{Krizhevsky2009a}
A. Krizhevsky.
\newblock {Learning Multiple Layers of Features from Tiny Images}.
\newblock \emph{University of Toronto}, 2009.

\bibitem[Li et~al.(2023)Li, Li, Zhao, Song, Li, and Yang]{Li2023a}
Z. Li, Y. Li, P. Zhao, R. Song, X. Li, and J. Yang.
\newblock {Is Synthetic Data from Diffusion Models Ready for Knowledge Distillation?}
\newblock \emph{{arXiv:2305.12954}}, 2023.

\bibitem[Liu et~al.(2024)Liu, Wang, Liu, Sun, and Yao]{Liu2024a}
H. Liu, Y. Wang, H. Liu, F. Sun, and A. Yao.
\newblock {Small Scale Data-Free Knowledge Distillation}.
\newblock In \emph{IEEE/CVF Conference on Computer Vision and Pattern Recognition}, 2024.

\bibitem[Liu et~al.(2022)Liu, Mao, Wu, Feichtenhofer, Darrell, and Xie]{Liu2022a}
Z. Liu, H. Mao, C.-Y. Wu, C. Feichtenhofer, T. Darrell, and S. Xie.
\newblock {A ConvNet for the 2020s}.
\newblock \emph{IEEE/CVF Conference on Computer Vision and Pattern Recognition}, 2022.

\bibitem[Maji et~al.(2013)Maji, Chicago, Rahtu, Kannala, Blaschk{\'{o}}, and Vedaldi]{Maji2013a}
S. Maji, T. Chicago, E. Rahtu, J. Kannala, M. Blaschk{\'{o}}, and A. Vedaldi.
\newblock {Fine-Grained Visual Classification of Aircraft}.
\newblock \emph{{arXiv preprint arxiv:1306.5151}}, 2013.

\bibitem[M{\"u}ller et~al.(2019)M{\"u}ller, Kornblith, and Hinton]{Muller2019a}
R. M{\"u}ller, S. Kornblith, and G.~E. Hinton.
\newblock {When Does Label Smoothing Help?}
\newblock \emph{Advances in Neural Information Processing Systems}, 2019.

\bibitem[Nayak et~al.(2019)Nayak, Mopuri, Shaj, Radhakrishnan, and Chakraborty]{Nayak2019a}
G.~K. Nayak, K.~R. Mopuri, V. Shaj, V.~B. Radhakrishnan, and A. Chakraborty.
\newblock {Zero-Shot Knowledge Distillation in Deep Networks}.
\newblock In \emph{International Conference on Machine Learning}, 2019.

\bibitem[Oquab et~al.(2023)Oquab, Darcet, Moutakanni, et~al.]{Oquab2023a}
M. Oquab, T. Darcet, T. Moutakanni, et~al.
\newblock {DinoV2: Learning Robust Visual Features Without Supervision}.
\newblock \emph{{arXiv:2304.07193}}, 2023.

\bibitem[Parkhi et~al.(2012)Parkhi, Vedaldi, Zisserman, and Jawahar]{Parkhi2012a}
O.~M. Parkhi, A. Vedaldi, A. Zisserman, and C.~V. Jawahar.
\newblock {Cats and Dogs}.
\newblock In \emph{IEEE Conference on Computer Vision and Pattern Recognition}, 2012.

\bibitem[Patel et~al.(2023)Patel, Mopuri, and Qiu]{Patel2023a}
G. Patel, K.~R. Mopuri, and Q. Qiu.
\newblock {Learning to Retain While Acquiring: Combating Distribution-Shift in Adversarial Data-Free Knowledge Distillation}.
\newblock In \emph{IEEE/CVF Conference on Computer Vision and Pattern Recognition}, 2023.

\bibitem[Radford et~al.(2021)Radford, Kim, Hallacy, et~al.]{Radford2021la}
A. Radford, J.~W. Kim, C. Hallacy, et~al.
\newblock {Learning Transferable Visual Models from Natural Language Supervision}.
\newblock In \emph{International Conference on Machine Learning}, 2021.

\bibitem[Romero et~al.(2015)Romero, Ballas, Kahou, Chassang, Gatta, and Bengio]{Romero2015a}
A. Romero, N. Ballas, S.~E. Kahou, A. Chassang, C. Gatta, and Y. Bengio.
\newblock {FitNets: Hints for Thin Deep Nets}.
\newblock In \emph{International Conference on Learning Representations}, 2015.

\bibitem[{S. Zagoruyko and N. Komodakis}(2016)]{Zagoruyko2016a}
{S. Zagoruyko and N. Komodakis}.
\newblock {Wide Residual Networks}.
\newblock In \emph{British Machine Vision Conference}, 2016.

\bibitem[Sandler et~al.(2018)Sandler, Howard, Zhu, Zhmoginov, and Chen]{Sandler2018a}
M. Sandler, A. Howard, M. Zhu, A. Zhmoginov, and L.-C. Chen.
\newblock {MobileNetV2: Inverted Residuals and Linear Bottlenecks}.
\newblock In \emph{IEEE/CVF Conference on Computer Vision and Pattern Recognition}, 2018.

\bibitem[Sarkar(1995)]{Sarkar1995a}
D. Sarkar.
\newblock {Methods to Speed Up Error Back-Propagation Learning Algorithm}.
\newblock \emph{ACM Computing Surveys}, 1995.

\bibitem[Shen and Savvides(2020)]{Shen2020a}
Z. Shen and M. Savvides.
\newblock {MEAL v2: Boosting Vanilla ResNet-50 to 80\%+ Top-1 Accuracy on ImageNet Without Tricks}.
\newblock In \emph{{Advances in Neural Information Processing Systems Workshops}}, 2020.

\bibitem[Shreiner et~al.(2009)]{Shreiner2009a}
D. Shreiner et~al.
\newblock \emph{{OpenGL Programming Guide: The Official Guide to Learning OpenGL, Versions 3.0 and 3.1}}.
\newblock {Pearson Education}, 2009.

\bibitem[Tang et~al.(2024{\natexlab{a}})Tang, Chen, Niu, Zhu, Zhou, Gong, and Sugiyama]{Tang2024b}
J. Tang, S. Chen, G. Niu, H. Zhu, J.~T. Zhou, C. Gong, and M. Sugiyama.
\newblock {Direct Distillation between Different Domains}.
\newblock \emph{arXiv:2401.06826}, 2024{\natexlab{a}}.

\bibitem[Tang et~al.(2022)Tang, Shakeel, Chen, Wan, and Kang]{Tang2022a}
W. Tang, M.~S. Shakeel, Z. Chen, H. Wan, and W. Kang.
\newblock {Target Category Agnostic Knowledge Distillation With Frequency-Domain Supervision}.
\newblock \emph{{IEEE Transactions on Industrial Informatics}}, 2022.

\bibitem[Tang et~al.(2024{\natexlab{b}})Tang, Zhang, Lv, Zhou, Duan, Kuang, and Wu]{Tang2024a}
Z. Tang, S. Zhang, Z. Lv, Y. Zhou, X. Duan, K. Kuang, and F. Wu.
\newblock {AuG-KD: Anchor-Based Mixup Generation for Out-of-Domain Knowledge Distillation}.
\newblock In \emph{International Conference on Learning Representations}, 2024{\natexlab{b}}.

\bibitem[Tarvainen and Valpola(2017)]{Tarvainen2017a}
A. Tarvainen and H. Valpola.
\newblock {Mean Teachers are Better Role Models: Weight-Averaged Consistency Targets Improve Semi-Supervised Deep Learning Results}.
\newblock \emph{Advances in Neural Information Processing Systems}, 2017.

\bibitem[Tian et~al.(2020)Tian, Krishnan, and Isola]{Tian2020a}
Y. Tian, D. Krishnan, and P. Isola.
\newblock {Contrastive Representation Distillation}.
\newblock In \emph{International Conference on Learning Representations}, 2020.

\bibitem[Tian et~al.(2023)Tian, Wang, Abdelmoniem, Liu, and Wang]{Tian2023a}
Z. Tian, Z. Wang, A.~M. Abdelmoniem, G. Liu, and C. Wang.
\newblock {Knowledge Representation of Training Data with Adversarial Examples Supporting Decision Boundary}.
\newblock \emph{{IEEE Transactions on Information Forensics and Security}}, 2023.

\bibitem[Touvron et~al.(2022)Touvron, Cord, and J{\'e}gou]{Touvron2022a}
H. Touvron, M. Cord, and H. J{\'e}gou.
\newblock {DeiT III: Revenge of the ViT}.
\newblock In \emph{European Conference on Computer Vision}, 2022.

\bibitem[Tran et~al.(2024)Tran, Le, Le, Harandi, Tran, and Phung]{Tran2024a}
M.-T. Tran, T. Le, X.-M. Le, M. Harandi, Q.~H. Tran, and D. Phung.
\newblock {NAYER: Noisy Layer Data Generation for Efficient and Effective Data-Free Knowledge Distillation}.
\newblock In \emph{IEEE/CVF Conference on Computer Vision and Pattern Recognition}, 2024.

\bibitem[TwiGL(2023)]{Twigl2023a}
TwiGL.
\newblock {TwiGL}.
\newblock \url{https://github.com/doxas/twigl}, 2023.
\newblock Accessed: June 29, 2023.

\bibitem[Urban et~al.(2017)Urban, Geras, Kahou, Aslan, Wang, Mohamed, Philipose, Richardson, and Caruana]{Urban2017a}
G. Urban, K.~J. Geras, S.~E. Kahou, O. Aslan, S. Wang, A. Mohamed, M. Philipose, M. Richardson, and R. Caruana.
\newblock {Do Deep Convolutional Nets Really Need to be Deep and Convolutional?}
\newblock In \emph{International Conference on Learning Representations}, 2017.

\bibitem[Wang(2021)]{Wang2021a}
Z. Wang.
\newblock {Data-Free Knowledge Distillation with Soft Targeted Transfer Set Synthesis}.
\newblock In \emph{AAAI Conference on Artificial Intelligence}, 2021.

\bibitem[Yin et~al.(2020)Yin, Molchanov, Alvarez, Li, Mallya, Hoiem, Jha, and Kautz]{Yin2020a}
H. Yin, P. Molchanov, J.~M. Alvarez, Z. Li, A. Mallya, D. Hoiem, N.~K. Jha, and J. Kautz.
\newblock {Dreaming to Distill: Data-Free Knowledge Transfer via DeepInversion}.
\newblock In \emph{IEEE/CVF Conference on Computer Vision and Pattern Recognition}, 2020.

\bibitem[Yoo et~al.(2019)Yoo, Cho, Kim, and Kang]{Yoo2019a}
J. Yoo, M. Cho, T. Kim, and U. Kang.
\newblock {Knowledge Extraction with No Observable Data}.
\newblock \emph{Advances in Neural Information Processing Systems}, 2019.

\bibitem[Yu et~al.(2023)Yu, Chen, Han, and Jiang]{Yu2023a}
S. Yu, J. Chen, H. Han, and s. Jiang.
\newblock {Data-Free Knowledge Distillation via Feature Exchange and Activation Region Constraint}.
\newblock In \emph{IEEE/CVF Conference on Computer Vision and Pattern Recognition}, 2023.

\bibitem[Zhang et~al.(2018)Zhang, Cisse, Dauphin, and Lopez-Paz]{Zhang2018a}
H. Zhang, M. Cisse, Y.~N. Dauphin, and D. Lopez-Paz.
\newblock {{\em mixup}: Beyond Empirical Risk Minimization}.
\newblock In \emph{International Conference on Learning Representations}, 2018.

\bibitem[Zhong et~al.(2020)Zhong, Zheng, Kang, Li, and Yang]{Zhong2020a}
Z. Zhong, L/ Zheng, G. Kang, S. Li, and Y. Yang.
\newblock {Random Erasing Data Augmentation}.
\newblock In \emph{AAAI Conference on Artificial Intelligence}, 2020.

\end{thebibliography}
}


\end{document}